\documentclass[format=acmsmall, review=false, screen=true]{acmart}

\usepackage[lined,ruled,linesnumbered]{algorithm2e}

\SetAlFnt{\small}
\SetAlCapFnt{\small}
\SetAlCapNameFnt{\small}
\SetAlCapHSkip{0pt}
\IncMargin{-\parindent}

\acmJournal{TOIS}
\acmYear{2019}
\copyrightyear{2019}

\setcopyright{acmlicensed}

\acmDOI{0000001.0000001}

\received{May 2018}
\received[accepted]{April 2019}

\newif\ifdraft
\drafttrue

\usepackage{amssymb}
\usepackage{amsmath}
\usepackage{array}
\usepackage[english]{babel}
\usepackage{color}
\usepackage{colortbl}
\usepackage{dcolumn}
\usepackage{graphicx}
\usepackage[utf8]{inputenc}
\usepackage{latexsym}
\usepackage{multicol}
\usepackage{multirow}
\usepackage{pifont}
\usepackage{rotating}
\usepackage{soul}
\usepackage{url}

\newcommand{\wasblue}[1]{\ifdraft{\leavevmode\color{black}{#1}}\else{\leavevmode\color{black}{#1}}\fi}

\newcommand{\fkfcv}{\textsc{Fun(kfcv)}}
\newcommand{\ftat}{\textsc{Fun(tat)}}

\newcommand{\sdag}{$^{\dagger}$}
\newcommand{\sddag}{$^{\dagger\dagger}$}
\newcommand{\sdpsdag}{$^{\dagger}$\phantom{$^{\dagger}$}}
\newcommand{\psdag}{\phantom{$^{\dagger}$}}
\newcommand{\psddag}{\phantom{$^{\dagger\dagger}$}}

\newtheorem{revcomment}{Reviewer Comment}[section]

\newcolumntype{M}[1]{>{\centering\arraybackslash}p{#1}}
\newcolumntype{L}{>{\centering\arraybackslash}m{12cm}}

\begin{document}
\title{Funnelling: A New Ensemble Method for Heterogeneous Transfer
Learning and its Application to Cross-Lingual Text Classification}
\thanks{The order in which the authors are listed is purely
alphabetical; each author has given an equally important contribution
to this work.}

\author{Andrea Esuli} \orcid{0000-0002-5725-4322} \affiliation{
\department{Istituto di Scienza e Tecnologie dell'Informazione}
\institution{Consiglio Nazionale delle Ricerche} \postcode{56124}
\city{Pisa} \country{Italy} } \email{andrea.esuli@isti.cnr.it}

\author{Alejandro Moreo} \orcid{0000-0002-0377-1025} \affiliation{
\department{Istituto di Scienza e Tecnologie dell'Informazione}
\institution{Consiglio Nazionale delle Ricerche} \postcode{56124}
\city{Pisa} \country{Italy} } \email{alejandro.moreo@isti.cnr.it}

\author{Fabrizio Sebastiani}\orcid{0000-0003-4221-6427} \affiliation{
\department{Istituto di Scienza e Tecnologie dell'Informazione}
\institution{Consiglio Nazionale delle Ricerche} \postcode{56124}
\city{Pisa} \country{Italy} } \email{fabrizio.sebastiani@isti.cnr.it}

\renewcommand{\shortauthors}{Esuli, Moreo, Sebastiani}

\renewcommand{\shorttitle}{Funnelling: A New Ensemble Method for
Heterogeneous Transfer Learning}


\begin{abstract}
  \emph{Cross-lingual Text Classification} (CLC) consists of
  automatically classifying, according to a common set $\mathcal{C}$
  of classes, documents each written in one of a set of languages
  $\mathcal{L}$, and doing so more accurately than when ``na\"ively''
  classifying each document via its corresponding language-specific
  classifier. In order to obtain an increase in the classification
  accuracy for a given language, the system thus needs to also
  leverage the training examples written in the other languages. We
  tackle ``multilabel'' CLC via \emph{funnelling}, a new ensemble
  learning method that we propose here. Funnelling consists of
  generating a two-tier classification system where all documents,
  irrespectively of language, are classified by the same (2nd-tier)
  classifier. For this classifier all documents are represented in a
  common, language-independent feature space consisting of the
  posterior probabilities generated by 1st-tier, language-dependent
  classifiers. This allows the classification of all test documents,
  of any language, to benefit from the information present in all
  training documents, of any language. We present substantial
  experiments, run on publicly available multilingual text
  collections, in which funnelling is shown to significantly
  outperform a number of state-of-the-art baselines. All code and
  datasets (in vector form) are made publicly available.
\end{abstract}



\maketitle


\section{Introduction}\label{sec:introduction}

\noindent In \emph{Multilingual Text Classification} (MLC) each
document $d$ is written in one of a finite set
$\mathcal{L}=\{\lambda_{1},...,$ $\lambda_{|\mathcal{L}|}\}$ of
languages, and the unlabelled documents need to be classified
according to a \emph{classification scheme}
$\mathcal{C}=\{c_{1},..., c_{|\mathcal{C}|}\}$ which is the same for
all $\lambda_{i}\in\mathcal{L}$.  \wasblue{MLC can be trivially solved as
$|\mathcal{L}|$ independent text classification tasks; in this case,
when MLC is solved via supervised learning, the training examples for
language $\lambda'$ have obviously no impact on the classifier for
language $\lambda''$. This is suboptimal, since it is somehow
intuitive that some cross-fertilization among the language-specific
classification tasks should be possible.

An important subtask of MLC that indeed tries to bring about this
cross-fertilization is \emph{Cross-Lingual Text Classification}
(CLC). In CLC set $\mathcal{L}$ is partitioned into a subset of
\emph{source languages} $\mathcal{L}^{s} \subset \mathcal{L}$ and a
subset of \emph{target languages}
$\mathcal{L}^{t}=\mathcal{L}/\mathcal{L}^{s}$; the goal is to build a
classifier $h_{i}$ for each target language
$\lambda_{i}\in\mathcal{L}^{t}$ despite the fact that a training set
$Tr_{i}$ for $\lambda_{i}$ might be too small, or might not exist at
all. CLC tries to accomplish this by leveraging the training data for
the source languages $\mathcal{L}^{s}$, for each of which a nonempty
training set of labelled documents is assumed available.
%
%
%

CLC is thus an instance of \emph{transfer learning}
\cite{Pan:2012fk,Vilalta:2011fk}, i.e., is a task in which we attempt
to reuse information about a problem in a source domain for solving
the same problem in a different, target domain. More specifically, CLC
is an instance of \emph{heterogeneous} transfer learning
\cite{Day:2017qe}, i.e., is a task in which transfer learning is
performed across domains that are characterized by different feature
spaces.  Techniques developed for CLC are especially useful when we
need to perform text classification for under-resourced languages,
i.e., languages for which only a small number (if at all) of training
documents are available; in these cases, CLC techniques allow
leveraging what is available for the better-resourced languages (e.g.,
English).

When a language $\lambda_{i}\in\mathcal{L}^{t}$ is such that no
training example exists for it, the task of CLC is to generate a
classifier for $\lambda_{i}$ that could not be generated
otherwise. This scenario is usually called \emph{zero-shot
cross-lingual classification}
(ZSCLC).}\footnote{\wasblue{\label{sec:terminology}The terminology in the
literature is, as in most fields of science, not entirely consistent;
in particular, what we here call ZSCLC is sometimes called CLC (see
e.g., \cite{Prettenhofer:2010ys,Moreo:2016fg}), and the scenario in
which training data are available for the target languages too is
sometimes called \emph{polylingual} TC (see e.g.,
\cite{Adeva:2005zx,Moreo:2016fk}). In this paper we have tried to
stick to the terminology that seems now the most widely adopted one.}}

Instead, when a language $\lambda_{i}\in\mathcal{L}^{t}$ is such that
a set $Tr_{i}$ of training documents is indeed available for it (which
is the scenario we will be mostly concerned with in this paper), so
that a (monolingual) classifier $h_{i}$ could in principle be
generated for $\lambda_{i}$, the task of CLC is to generate an
``enhanced'' classifier $h^{+}_{i}$ (i.e., a classifier more accurate
than $h_{i}$) by also leveraging the training examples in
$\mathcal{L}^{s}$. Note that, when training data are available for
each $\lambda_{i}\in\mathcal{L}$, each $\lambda_{i}$ can alternatively
play the role of the source or of the target language, i.e.,
unlabelled data in any language can benefit from the training data in
any language.

In this paper we will focus on general \emph{multilabel} CLC, i.e.,
the CLC case in which the number of classes to which a document $d$
belongs ranges in $\{0, ..., |\mathcal{C}|\}$; note that multilabel
CLC subsumes binary classification (which corresponds to multilabel
CLC with $|\mathcal{C}|=1$).  We propose a new, learner-independent
approach for multilabel CLC that relies on \emph{funnelling}, a 2-tier
method for training classifier ensembles for heterogeneous data (i.e.,
data that lie in different feature spaces), which is being proposed
here for the first time. In our approach a test document $d_{u}$
written in language $\lambda_{i}$ is classified by $h^{1}_{i}$, one
among $|\mathcal{L}|$ language-specific multilabel \emph{base
classifiers}, and the output of this classifier (in the form of a
vector of $|\mathcal{C}|$ posterior probabilities $\Pr(c|d_{u})$) is
input to a multilabel \emph{meta-classifier} which generates the final
prediction for $d_{u}$ using the latter vector as $d_{u}$'s
representation.

The base classifiers can actually be seen as mapping $|\mathcal{L}|$
different language-dependent feature spaces $\phi^{1}_{i}$ (e.g.,
consisting of terms or other content features) into a common,
language-independent feature space $\phi^{2}$ (consisting of posterior
probabilities). In other words, documents written in different
languages, that in the 1st tier lie in different feature spaces, in
the 2nd tier are ``funnelled'' into a single feature space.
One advantage of this fact is that, as will become clear in Section
\ref{sec:funnelling}, \emph{all} training examples (irrespectively of
language) contribute to training the meta-classifier. As a result, the
classification of unlabelled documents written in \emph{any} of the
languages in $\mathcal{L}$ benefits from \emph{all} the training
examples, written in any language of $\mathcal{L}$, and thus delivers
better results. Another advantage of this approach to CLC is its
complete generality, since funnelling does not require the
availability of multilingual dictionaries, machine translation
services, or external corpora (either parallel or comparable).

This paper is structured as follows. After some discussion of related
work (Section \ref{sec:relatedwork}), in Section \ref{sec:funnelling}
we describe our approach to multilabel CLC in detail; in particular,
in Section \ref{sec:funnellingandstacking} we take a critical look at
funnelling and at its relationships with \emph{stacked generalization}
\cite{Wolpert:1992rq}, and we discuss what exactly one attempts to
learn via funnelling. In Sections \ref{sec:experimentalsetting} and
\ref{sec:results} we turn to describing the substantive
experimentation to which we subject our approach; in particular, we
describe experiments in multilabel CLC settings (Section
\ref{sec:MLCLCresults}),
in monolingual settings and in binary settings (Section
\ref{sec:monolingualresults}), and in settings that aim to show how
funnelling may help classification for under-resourced languages
(Sections \ref{sec:lessresourced} and \ref{sec:languages}). \wasblue{In
this paper we mostly focus on the situation in which some training
data are indeed available also for each of the target languages;
Section \ref{sec:ZSCLC} is instead devoted to discussing how
funnelling can be adapted to the zero-shot case.}  Section
\ref{sec:conclusion} concludes, pointing at possible avenues for
future work.


\section{Related work}\label{sec:relatedwork}

\noindent 
Initial work on CLC \cite{Bel03,Adeva:2005zx} relied on standard
bag-of-words representations, and investigated different preprocessing
techniques with simple strategies for classification based on
language-specific feature spaces (giving rise to one classifier for
each language) or a single juxtaposed feature space (giving rise to
one single classifier for the entire set of languages).  Since then,
more sophisticated \emph{distributional semantic models} (DSMs), such
as \emph{Cross-Lingual Latent Semantic Analysis} (CLLSA --
\cite{dumais1997automatic}) and \emph{Polylingual Latent Dirichlet
Allocation} (PLDA -- \cite{Mimno:2009hp}), have been extensively
investigated. 
However, the improvement in accuracy brought about by models based on
these latent representations comes at a cost, since the availability
of external parallel corpora (i.e., additional to the one used for
training and testing purposes) is typically required.

In the absence of external parallel data, one cross-lingual DSM which
has recently proved worthy (and that we use as a baseline in our
experiments) is \emph{Lightweight Random Indexing} (LRI --
\cite{Moreo:2016fk}), the multilingual extension of the \emph{Random
Indexing} (RI) method 
\cite{Sahlgren:2004ph}. RI is a context-counting model belonging to
the family of random projection methods, and is considered a cheaper
approximation of LSA \cite{sahlgren2005introduction}. LRI is designed
so that the orthogonality of the projection base is maximized, which
allows to preserve sparsity and maximize the contribution of the
information conveyed by the features shared across languages.


Other techniques (e.g., \cite{Franco-Salvador:2014ve})
rely, in order to solve the multilingual classification problem, on
the availability of external multilingual knowledge resources, such as
dictionaries or thesauri.
One of the best-known such approaches (which we will also use as a
baseline in our experiments) is \emph{Cross-Lingual Explicit Semantic
Analysis} (CLESA -- \cite{Song:2016fr,Sorg:2008qv}). In the original
monolingual version of this technique a document is represented by a
vector of similarity values, where each such value represents the
similarity between the document and a predefined reference text
\cite{Gabrilovich:2007ss}. In CLESA, different language-specific
versions of the same text are considered as reference texts, so that
documents written in different languages can be effectively
represented in the same feature space.  \wasblue{In a similar vein,
\emph{Kernel Canonical Correlation Analysis} (KCCA)
\cite{hardoon2004canonical}, the kernelized version of CCA
\cite{hotelling1936relations}, has also been applied to cross-lingual
contexts.
In essence, CCA aims at maximizing the correlations among sets of
variables via linear projections onto a shared space.  In its
application to cross-lingual classification, KCCA (which we will also
use as a baseline in our experiments) treats language-specific views
of aligned articles as different sets of variables to correlate.
The projections that maximize the correlations among language-specific
aligned articles are applied to the training documents in order to
create a classifier.}

Another method that requires external multilingual resources
(specifically: a word translation oracle) is \emph{Cross-Lingual
Structural Correspondence Learning} (CL-SCL --
\cite{Prettenhofer:2010ys}).  CL-SCL relies on solving auxiliary
prediction problems, which consist in discovering hidden correlations
between terms in a language.  This is achieved by binary classifiers
trained to predict the presence of highly discriminative terms
(``pivots'') given the other terms in the document.  The cross-lingual
aspect is addressed by imposing that pivot terms are aligned (i.e.,
translations of each other) across languages, which requires a word
translation oracle.  A stronger, more recent variant of CL-SCL (which
we also compare against in our experiments) is \emph{Distributional
Correspondence Indexing} (DCI -- \cite{Moreo:2016fg}).  DCI derives
term representations in a vector space common to all languages where
each dimension reflects its distributional correspondence (as
quantified by a ``distributional correspondence function'') to a
pivot.
  
Machine Translation (MT) represents an appealing tool to solve CLC,
and several CLC methods are indeed based on the use of MT services
\cite{Rigutini:2005bk,wan2009co}. However, the drawback of these
methods is reduced generality, since it is not always the case that
quality MT tools are both (a) available for the required language
combinations, and (b) free to use.

Approaches to CLC based on deep learning focus on defining
representations based on word embeddings which capture the semantic
regularities in language while at the same time being aligned across
languages. In order to produce aligned representations, though, deep
learning approaches typically require the availability of external
parallel corpora \cite{Klementiev:2012pi,Gouws:2015zm},
\wasblue{bi-lingual dictionaries \cite{mikolov2013exploiting}},
bi-lingual lexicons \cite{Faruqui:2014fk}, or machine translation
tools
\cite{Balikas:2016lk}. 
Recently, \citet{Conneau:2018bv} proposed a method to align
monolingual word embedding spaces (as those produced by, e.g.,
Word2Vec \cite{Mikolov:2013jk}) from different languages without
requiring parallel data. To this aim, \cite{Conneau:2018bv} proposed
an adversarial training process in which a \emph{generator} (in charge
of mapping the source embeddings onto the target space) is trained to
fool a \emph{discriminator} from distinguishing the provenance of the
embeddings, i.e., from understanding whether the embeddings it
receives as input come from the (transformed) source or from the
target space. After that, the mapping is refined by means of
unsupervised techniques. Despite operating without parallel resources,
\cite{Conneau:2018bv} obtained state-of-the-art multilingual mappings,
which they later made publicly
available\footnote{\url{https://github.com/facebookresearch/MUSE}} and
which we use as a further baseline in our experiments of Section
\ref{sec:experimentalsetting}. \wasblue{We refer the interested reader to
\cite{Ruder:2017wj} for a comprehensive survey on the most important
techniques for generating multilingual embeddings, and to
\cite{Upadhyay:2016dq} for an empirical comparison of different such
techniques on several cross-lingual tasks.}

\section{Solving cross-lingual text classification via
funnelling}\label{sec:funnelling}



\noindent
We now describe funnelling and its application to multilabel CLC. Let
$\mathcal{L}=\{\lambda_{1},...,\lambda_{|\mathcal{L}|}\}$ be our
finite set of languages, and let
$\mathcal{C}=\{c_{1},..., c_{|\mathcal{C}|}\}$ be our finite
classification scheme. Let $d$ indicate a generic document, $d_{l}$ a
labelled (training) document, and $d_{u}$ an unlabelled (test)
document. We assume the existence of $|\mathcal{L}|$ \wasblue{nonempty}
training sets $\{Tr_{1},...,Tr_{|\mathcal{L}|}\}$ of documents, where
all documents $d_{l} \in Tr_{i}$ are written in language $\lambda_{i}$
and are labelled according to $\mathcal{C}$ (i.e., the set
$\mathcal{C}$ of classes is the same for all training sets). We do not
make any assumption on the relative size and composition of the
different training sets; we thus allow different training sets to
consist of different numbers of documents, and we do not assume the
union of the training sets to be either a ``parallel'' dataset (i.e.,
consisting of translation-equivalent versions of the same documents)
or a ``comparable'' one (i.e., consisting of documents dealing with
the same events/topics although in different languages).

The first step of the training process consists of training
$|\mathcal{L}|$ independent base classifiers
$h^{1}_{1},..., h^{1}_{|\mathcal{L}|}$ from the respective training
sets (throughout this paper the ``1'' superscript will indicate the
1st tier of the architecture, which consists of the base
classifiers). In order to do this, for each training document
$d_{l}\in Tr_{i}$ we generate a vectorial representation
$\phi^{1}_{i}(d_{l})$ via bag-of-words or any other standard
content-based representation model; we use all the resulting vectors
to train $h^{1}_{i}$, and repeat the process for all the
$Tr_{i}$'s. Quite obviously, the different base classifiers will
operate in different feature spaces \wasblue{(for a detailed
discussion of this point see the last paragraph of Section
\ref{sec:representingtext})}.

We do not make any assumption concerning (a) the model used for
generating the vectorial representations $\phi^{1}_{i}(d)$ and (b)
the supervised learning algorithm used to train the base classifiers;
it is in principle possible to use different representation models and
different supervised learning algorithms for the different
languages. Actually, the only assumption we make is that each trained
base classifier $h^{1}_{i}$ returns, for each document $d_{u}$ written
in language $\lambda_{i}$ and for each class $c$, a classification
score $h^{1}_{i}(d_{u},c)\in\mathbb{R}$, i.e., a numerical value
representing the confidence that $h^{1}_{i}$ has in the fact that
$d_{u}$ belongs to
$c$. 

The second step consists of generating, for each document
$d_{l}\in Tr_{i}$ and for each training set $Tr_{i}$, a vectorial
representation $\phi^{2}(d_{l})$ that will be used for training the
meta-classifier. In order to do this, for each document
$d_{l}\in Tr_{i}$ we first generate a vector
\begin{equation}
  \begin{aligned}
    \label{eq:Sd}
    S(d_{l}) =(h^{1}_{ix}(d_{l},c_{1}),...,
    h^{1}_{ix}(d_{l},c_{|\mathcal{C}|}))
  \end{aligned}
\end{equation}
\noindent of \wasblue{$|\mathcal{C}|$} classification scores,
\wasblue{one per class,} via $k$-fold cross-validation on $Tr_{i}$. In
other words, we split $Tr_{i}$ into $k$ subsets $Tr_{i1},..., Tr_{ik}$
of approximately equal size, train a classifier $h^{1}_{ix}$ (using
$\phi^{1}_{i}(d)$-style vectorial representations for the training
documents) using the training data in
$\bigcup_{y\in \{1, ..., k\}, y\not = x}Tr_{iy}$, use this classifier
in order to generate vectors $S(d_{l})$ of classification scores for
all $d_{l}\in Tr_{ix}$, and repeat the process for all
$1\leq x\leq k$. \wasblue{The reason why we use $k$-fold
cross-validation is that we want the classification scores which
vector $S(d_{l})$ is composed of, to be generated by classifiers
trained on data that do not contain $d_{l}$ itself.}


All training documents, irrespectively of the language they are
written in, thus give rise to (dense) vectors $S(d_{l})$ of
classification scores, \wasblue{and these vectors are all in the same
vector space}. In other words, should we view a document $d_{l}$ as
represented by vector $S(d_{l})$, all documents would be represented
in the same feature space, i.e., the space of base classifier scores
for classes $\mathcal{C}=\{c_{1},..., c_{|\mathcal{C}|}\}$. We could
thus in principle use the set
$\{S(d_{l}) \ | \ d_{l}\in\bigcup_{i=1}^{|\mathcal{L}|}Tr_{i}\}$ as a
large unified training set for training a meta-classifier for
$\mathcal{C}$.  This is indeed what we are going to do, but before
doing this we transform all vectors $S(d_{l})$ of classification
scores into vectors of \wasblue{$|\mathcal{C}|$} posterior
probabilities
\begin{equation}
  \begin{aligned}
    \label{eq:phi2d}
    \phi^{2}(d_{l})= & \ (\Pr(c_{1}|d_{l}),...,
    \Pr(c_{|\mathcal{C}|}|d_{l})) \\ = & \
    (f_{ix}(h^{1}_{ix}(d_{l},c_{1})),...,
    f_{ix}(h^{1}_{ix}(d_{l},c_{|\mathcal{C}|})))
  \end{aligned}
\end{equation}
\noindent where $\Pr(c_{j}|d_{l})$ represents the probability that the
originating base classifier attributes to the fact that $d_{l}$
belongs to $c_{j}$, and where $f_{ix}$ is a mapping to be discussed
shortly. Note that the $\Pr(c_{j}|d_{l})$'s are just subjective
estimates generated by the classifiers, and are not probabilities in
any ``objective'' sense (whatever this might mean).

The rationale for not using the original classification scores
$h^{1}_{ix}(d_{l},c_{j})$ as features is that vectors of
classification scores coming from different classifiers are not
comparable with each other (see \cite[\S 7.1.3]{Bishop:2006mw} for a
discussion),
and it would thus be unsuitable to use them \emph{together} as feature
vectors in the same training set. The task of finding a function
$f_{ix}$ that maps classification scores into posterior probabilities
while at the same time obtaining ``well calibrated'' (i.e., good)
posterior probabilities, is referred to as \emph{probability
calibration},\footnote{Posterior probabilities $\Pr(c|d)$ are said to
be \emph{well calibrated} when
$\lim_{|S|\rightarrow \infty}\frac{|\{d\in c| \Pr(c|d)=x\}|}{|\{d\in
S| \Pr(c|d)=x\}|}=x$ \cite{DeGroot:1983pt}. Intuitively, this property
implies that, as the size of the sample $S$ goes to infinity, e.g.,
90\% of the documents $d\in S$ such that $\Pr(c|d)=0.9$ belong to
class $c$. Some learning algorithms (e.g., AdaBoost, SVMs) generate
classifiers that return confidence scores that are not probabilities,
since these scores do not range on [0,1]; in this case, a calibration
phase is needed to convert these scores into well calibrated
probabilities. Other learning algorithms (e.g., Na\"ive Bayes)
generate classifiers that output probabilities that are not well
calibrated; in this case too, a calibration phase is necessary in
order to obtain well calibrated probabilities. Yet other learning
algorithms (e.g., logistic regression) are known to generate
classifiers that already return well calibrated probabilities; in
these cases no separate calibration phase is necessary.}
and several methods for performing it are known from the literature
(see e.g., \cite{Platt:2000fk,Wu:2004rz}).  We perform probability
calibration independently for each of the $|\mathcal{L}|$ training
sets and each of the $k$ folds (since each of these
$|\mathcal{L}|\times k$ settings yields a different classifier), thus
resulting in $|\mathcal{L}|\times k$ different calibration functions
$f_{11}, ..., f_{|\mathcal{L}|k}$.

The net result is that all the vectors in
$\{\phi^{2}(d_{l})|d_{l}\in \bigcup_{i=1}^{|\mathcal{L}|}Tr_{i}\}$ are
now comparable, and can thus be safely used for training the
meta-classifier $h^{2}$. Here we do not make any assumption concerning
the learning algorithm used to train $h^{2}$, the only requirement
being that it needs to accept non-binary vectorial representations as
input.  In particular, it is in principle possible to train our
meta-classifier via a learning algorithm different from the one used
to train the base classifiers.

As a final step of the learning process we perform probability
calibration for the base classifiers
$h^{1}_{1},..., h^{1}_{|\mathcal{L}|}$ trained in the first step, thus
giving rise to additional $|\mathcal{L}|$ calibration functions
$f_{1},..., f_{|\mathcal{L}|}$.

The classification process follows the steps already outlined in
Section \ref{sec:introduction}. An unlabelled document $d_{u}$ written
in language $\lambda_{i}\in \mathcal{L}$ is classified by its
corresponding language-specific base classifier $h^{1}_{i}$. The
resulting vector of classification scores $S(d_{u})$ is mapped into a
vector $\phi^{2}(d_{u})$ of posterior probabilities by the function
$f_{i}$ obtained via probability calibration in the last step of the
training process. Vector $\phi^{2}(d_{u})$ is fed to classifier
$h^{2}$,
which generates $|\mathcal{C}|$ binary classification decisions
$h^{2}(d_{u},c_{1}),...,$ $h^{2}(d_{u},c_{|\mathcal{C}|})$.

We call our method \fkfcv\ -- with \textsc{kfcv} standing for
``$k$-Fold Cross-Validation'' -- in order to distinguish it from a
variant to be discussed in Section \ref{sec:twovariantsoffunnelling}.


\subsection{Two variants of funnelling}
\label{sec:twovariantsoffunnelling}

\noindent One problem with \fkfcv\ is that the representations
$\phi^{2}(d_{l})$ of the labelled documents used to train the
meta-classifier $h^{2}$ may not match well (i.e., faithfully
represent) the representations $\phi^{2}(d_{u})$ of the unlabelled
documents that will be fed to $h^{2}$, and this would contradict the
basic assumption of supervised learning. In fact, (assuming for
simplicity that both $d_{l}$ and $d_{u}$ are written in the same
language $\lambda_{i}$) the posterior probabilities of which
$\phi^{2}(d_{u})$ consists of have been generated by classifier
$h^{1}_{i}$, which has been trained on the entire set $Tr_{i}$, while
the posterior probabilities of which $\phi^{2}(d_{l})$ consists of,
have been generated by one of the classifiers $h^{1}_{ix}$ trained
during the $k$-fold cross-validation process, which has been trained
on a \emph{subset} of $Tr_{i}$ of cardinality $\frac{k-1}{k}|Tr_{i}|$.

In other words, the base classifier $h^{1}_{i}$ that classifies the
unlabelled documents has received \emph{more} training than the base
classifiers $h^{1}_{ix}$ that classified the training data; this
difference may be especially substantial for low-frequency classes,
where decreasing the size of the training set sometimes means
depleting an already tiny set of positive training examples. As a
result, the posterior probabilities $\Pr(c_{j}|d_{u})$ for the
unlabelled documents tend to be different (actually: higher-quality)
than the corresponding posterior probabilities $\Pr(c_{j}|d_{l})$ for
the training documents. Because of this mismatch, the meta-classifier
$h^{2}$ may perform suboptimally.

In order to minimize this mismatch one could arbitrarily increase the
number $k$ of folds, maybe even using leave-one-out validation (i.e.,
$k$-fold cross-validation with $k=|Tr_{i}|$). However, this solution
is computationally impractical, since a high value of $k$ implies not
only a high number of training rounds, but also a high number of
probability calibration rounds (since, as already observed,
calibration needs to be done independently for each trained
classifier), which is expensive since calibration usually entails
extensive search in a space of parameters.

An alternative, radically simpler solution might consist in doing away
with $k$-fold cross-validation. In this solution (that we will call
\ftat, where \textsc{tat} stands for ``Train and Test''), Equations
\ref{eq:Sd} and \ref{eq:phi2d} would be replaced by
\begin{align}
  \label{eq:Sdnokfcv}
  S(d_{l}) = & \ (h^{1}_{i}(d_{l},c_{1}), ..., h^{1}_{i}(d_{l},c_{|\mathcal{C}|})) \\
  \phi^{2}(d_{l}) = & \ (\Pr(c_{1}|d_{l}),...,
                      \Pr(c_{|\mathcal{C}|}|d_{l})) \\ = 
             & \ (f_{i}(h^{1}_{i}(d_{l},c_{1})),...,
               f_{i}(h^{1}_{i}(d_{l},c_{|\mathcal{C}|})))
\end{align}
\noindent i.e., the vectors of \wasblue{$|\mathcal{C}|$} scores
$S(d_{l})$ and the vectors $\phi^{2}(d_{l}) $ of
\wasblue{$|\mathcal{C}|$} posterior probabilities would be generated
directly by the classifiers $h^{1}_{i}$ trained on the entire training
set $Tr_{i}$ (with the help of the calibration functions $f_{i}$
discussed towards the end of the previous section). Note that \ftat\
entails just $|\mathcal{L}|$ training and calibrations rounds, while
\fkfcv\ entails $|\mathcal{L}|\times (k+1)$.

\ftat\ is not exempt from problems either, and actually suffers from
the \emph{opposite} drawback with respect to \fkfcv. Here again, the
representations $\phi^{2}(d_{l})$ of the labelled documents used to
train the meta-classifier may not match well the representations
$\phi^{2}(d_{u})$ of the unlabelled documents, for the simple reason
that
classifier $h^{1}_{i}$ classifies (in order to generate the
representations $\phi^{2}(d_{l})$ to be used for training the
meta-classifier) the very same training examples $d_{l}$ it has been
trained on. As a result, the posterior probabilities
$\Pr(c_{j}|d_{u})$ for the unlabelled documents tend to be
\emph{lower-} quality (hence different) than the corresponding
posterior probabilities $\Pr(c_{j}|d_{l})$ for the training documents,
since documents $d_{u}$ have not been seen during training.

The two variants have thus opposite pros and cons; as a result, in our
experiments we will test both of them, side by side. Both variants are
collectively described in pseudocode form as Algorithm \ref{alg:fun},
where the \textbf{if} command of Line \ref{step:whichvariant}
determines which of the two variants is executed.

\begin{algorithm}
  \SetNoFillComment
  \begin{footnotesize}
    \SetKwInOut{Input}{Input} \SetKwInOut{Output}{Output}
    \Input{\textbullet\ Sets $\{Tr_{1},...,Tr_{|\mathcal{L}|}\}$ of
    training documents written in
    languages
    $\mathcal{L}=\{\lambda_{1},...,\lambda_{|\mathcal{L}|}\}$, all
    labelled according \\
    \hspace{.6em} to sets of
    classes
    $\mathcal{C}=\{c_{1},..., c_{|\mathcal{C}|}\}$; \\
    \hspace{0em} \textbullet\ Sets $\{Te_{1},...,Te_{|\mathcal{L}|}\}$
    of unlabelled documents written in
    languages
    $\mathcal{L}=\{\lambda_{1},...,\lambda_{|\mathcal{L}|}\}$, all to
    be labelled \\
    \hspace{.6em} according to sets of
    classes
    $\mathcal{C}=\{c_{1},..., c_{|\mathcal{C}|}\}$; \\
    \hspace{0em} \textbullet\ Flag \emph{Variant}, with values in
    \{\emph{\fkfcv}, \emph{\ftat}\} } \Output{\textbullet\ 1st-tier
    language-specific classifiers
    $h^{1}_{1},..., h^{1}_{|\mathcal{L}|}$ ; \\
    \hspace{0em} \textbullet\ 2nd-tier language-independent classifier $h^{2}$ ; \\
    \hspace{0em} \textbullet\ Labels for all documents in
    $\{Te_{1},...,Te_{|\mathcal{L}|}\}$;} \BlankLine
    \tcc{Training phase} \For{$\lambda_{i}\in \mathcal{L}$}{
    \tcc{Train 1st-tier classifiers and find a calibration function
    for them}
    Train classifier $h^{1}_{i}$ from $Tr_{i}$; \\
    Compute calibration function $f_{i}$ via chosen calibration method; \\
    \tcc{Generate vectors of posterior probabilities for training
    meta-classifiers}
    \eIf{Variant=``\fkfcv''}{\label{step:whichvariant} \tcc{Use the
    \fkfcv\ variant of the algorithm}
    Split $Tr_{i}$ into $k$ folds $\{Tr_{i1}$,...,$Tr_{ik}\}$; \\
    \For{$1\leq x\leq k$}{
    Train classifier $h^{1}_{ix}$ from $\bigcup_{y\in \{1, ..., k\}, y\not = x}Tr_{iy}$; \\
    Compute calibration function $f_{ix}$ via chosen calibration method; \\
    \For{$d_{l}\in Tr_{ix}$}{ \tcc{Compute vector of calibrated
    posterior probabilities}
    $\phi^{2}(d_{l})\leftarrow(f_{ix}(h^{1}_{ix}(d_{l},c_{1})), ... , f_{ix}(h^{1}_{ix}(d_{l},c_{|\mathcal{C}|})))$  \label{step:postprob1} ; \\
    }}}{ \tcc{Use the \ftat\ variant of the algorithm}
    \For{$d_{l}\in Tr_{i}$}{ \tcc{Compute vector of calibrated
    posterior probabilities}
    $\phi^{2}(d_{l})\leftarrow(f_{i}(h^{1}_{i}(d_{l},c_{1})),..., f_{i}(h^{1}_{i}(d_{l},c_{|\mathcal{C}|})))$  \label{step:postprob2} ; \\
    } } } \BlankLine
    Train classifier $h^{2}$ from all vectors $\phi^{2}(d_{l})$;
    \BlankLine
    \tcc{Classification phase} \For{$\lambda_{i}\in \mathcal{L}$}{
    \For{$d_{u}\in Te_{i}$}{ \tcc{Compute vector of calibrated
    posterior probabilities}
    $\phi^{2}(d_{u})\leftarrow (f_{i}(h^{1}_{i}(d_{u},c_{1})), ..., f_{i}(h^{1}_{i}(d_{u},c_{|\mathcal{C}|})))$ ; \\
    \tcc{Invoke meta-classifier}
    Compute $h^{2}(d_{u},c_{1}),...,$ $h^{2}(d_{u},c_{|\mathcal{C}|})$ from $\phi^{2}(d_{u})$.  \label{step:finalstep1}\\
    }}
    \caption{Funnelling for multilabel CLC; the
    \textbf{if} command of Line \ref{step:whichvariant} chooses which
    of \fkfcv\ and \ftat\ is executed. 
    }\label{alg:fun}
  \end{footnotesize}
\end{algorithm}\DecMargin{1em}



\subsection{What does funnelling learn,
exactly?} \label{sec:funnellingandstacking}

\noindent
Funnelling is reminiscent of the \emph{stacked generalization}
(a.k.a.\ ``stacking'') method for ensemble learning
\cite{Wolpert:1992rq}.
Let us discuss their commonalities and differences.

Common to stacking and funnelling is the presence of an ensemble of
$n$ base classifiers, typically trained on ``traditional'' vectorial
representations, and the presence of a single meta-classifier that
operates on vectors of base-classifier outputs. Common to stacking and
\fkfcv\ is also the use of $k$-fold cross-validation in order to
generate the vectors of base-classifier outputs that are used to train
the meta-classifier. (Variants of stacking in which $k$-fold
cross-validation is not used, and thus akin to \ftat, also exist
\cite{Sakkis:2001ms}.)

However, a key difference between the two methods is that stacking
(like other ensemble methods such as bagging \cite{Breiman:1996kx} and
boosting \cite{Freund:1996wd}) deals with (``homogeneous'') scenarios
in which all training documents can in principle be represented in the
same feature space and can thus concur to training the same
classifier; in turn, this classifier can be used for classifying all
the unlabelled documents. In stacking, the base classifiers sometimes
differ in terms of the learning algorithm used to train them
\cite{Sakkis:2001ms,Ting:1999si},
or in terms of the subsets of the training set which are used for
training them \cite{Chan:1997uq}.  In other words, in these scenarios
setting up an ensemble is a choice, and not a necessity. It is instead
a necessity in the (``heterogeneous'') scenarios which funnelling
deals with, where labelled documents of different types (in our case:
languages) could otherwise \emph{not} concur in training the same
classifier (since they lie in different feature spaces), and where
unlabelled documents could not (for analogous reasons) be classified
by the same classifier.

\begin{figure}[tb]
  \centering
  \includegraphics[width=.9\textwidth]{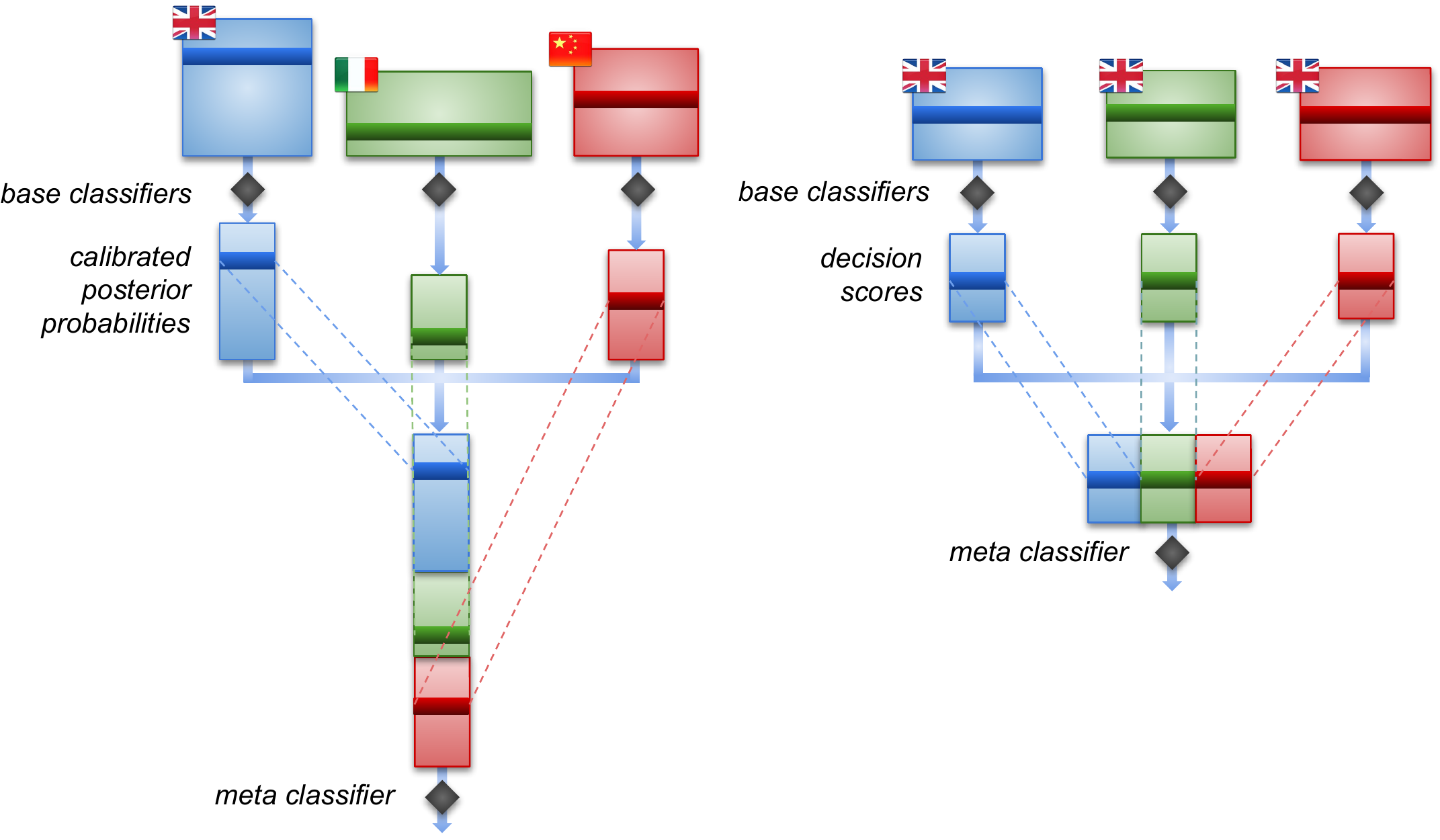}
  \caption{Architectures of a classifier system based on funnelling
  (left) and of one based on stacking (right). Black diamonds
  represent individual classifiers, dark thin coloured rectangles
  represent individual vectors, while larger coloured rectangles that
  contain them represent sets of vectors; national flags represent the
  different languages on which language-specific classifiers
  operate. The fact that, in funnelling, the larger coloured
  rectangles at the top have different widths indicates that the sets
  of vectors they represent lie in different feature spaces, which may
  have different dimensionalities (this is usually not the case in
  stacking); the fact that they have different heights indicates that
  the sets of vectors they represent may come in different sizes (this
  is usually not the case in stacking either); above all, the fact
  that they are labelled by different national flags indicates that
  the sets of vectors they represent lie in different feature spaces.}
  \label{fig:architecture}
\end{figure}

The consequence is that, while in stacking \emph{all} base classifiers
classify the test document, in funnelling only one base classifier
does this.\footnote{\label{foot:Kunch}\wasblue{Kuncheva \cite[p.\
106]{Kuncheva:2004mz} observes that ``It is accepted now that there
are two main strategies in combining classifiers: fusion and
selection. In classifier fusion, each ensemble member is supposed to
have knowledge of the whole feature space. In classifier selection,
each ensemble member is supposed to know well a part of the feature
space and be responsible for objects in this part.'' Funnelling is
thus an instance of the ``classifier selection'' strategy for creating
an ensemble.}} In turn, this means that in stacked generalization the
length of the vectors on which the meta-classifier operates is
$n\cdot |\mathcal{C}|$ (with $n$ the number of base classifiers),
while it is just $|\mathcal{C}|$ in funnelling. In stacking, $n$
different scores (one for each base classifier) for the same
$(d_{u},c)$ test pair are thus received by the meta-classifier, who
then needs to combine them in order to reach a final decision. As
noted in \cite{Dzeroski:2004qh}, stacking is indeed a method for
\emph{learning to combine} the $n$ scores returned by a set of $n$
base classifiers for the same $(d_{u},c)$ test pair. While in many
classifier ensembles a static combination rule -- e.g., weighted
voting -- is used to combine the outputs of the individual base
classifiers, in stacking this combination rule is learned from
data. By contrast, there is no combination of different outputs in
funnelling, since a document is always classified by only one base
classifier.  Graphical depictions of the architectures of funnelling
and stacking are given in Figure \ref{fig:architecture}.

So, if the meta-classifier of an ensemble built via funnelling does
not learn to combine different scores for the same $(d_{u},c)$ pair,
what does it learn exactly?

It certainly \emph{learns to exploit the stochastic dependencies
between classes} that exist in multilabel settings
\cite{Godbole:2004eu,Ness:2009th,Tsoumakas:2007fx}, which is not
possible when (as customarily done) a multilabel classification task
is solved as $|\mathcal{C}|$ independent binary classification
problems. In fact, for an unlabelled document $d_{u}$ the
meta-classifier receives $|\mathcal{C}|$ inputs from the base
classifier which has classified $d_{u}$, and returns $|\mathcal{C}|$
outputs, which means that the input for class $c'$ has a potential
impact on the output for class $c''$, for every choice of $c'$ and
$c''$. For instance, the fact that for $d_{u}$ the posterior
probability for class \texttt{Skiing} is high might bring additional
evidence that $d_{u}$ belongs to class \texttt{Snowboarding}; this
could be the result of several training documents labelled by
\texttt{Snowboarding} having, in their $\phi^{2}(d)$ vectors, a high
value for class \texttt{Skiing}.



However, learning to exploit the stochastic dependencies between
different classes
is certainly not the primary motivation behind funnelling. The primary
motivation is instead \emph{learning from heterogeneous data}, i.e.,
data that come in $n$ different, incomparable varieties, and that
because of the differences among these varieties require $n$
completely different feature spaces to accommodate them. When all
these diverse data need to be classified, despite their diversity,
according to a common classification scheme $\mathcal{C}$, funnelling
can be used to set up a single classifier (the meta-classifier) that
handles them all. Funnelling can be seen as mapping $n$ different,
incomparable feature spaces into a common, more abstract feature space
in which all differences among the original $n$ feature spaces have
been factored out. As a result, the meta-classifier can be trained
from the union of the $n$ training sets, which means that \emph{all}
training examples, irrespectively of their provenance, concur to the
common goal of classifying \emph{all} the unlabelled examples,
irrespectively of the provenance of each of these.




\section{Experimental setting}\label{sec:experimentalsetting}


\subsection{Datasets}\label{sec:dataset}

\noindent We perform our experiments on two publicly available
datasets, RCV1/RCV2 (a comparable corpus) and JRC-Acquis (a parallel
corpus).\footnote{All the information required to replicate the
experiments, e.g., IDs of the selected documents, assigned labels,
code, etc., is made available at
\url{https://github.com/AlexMoreo/funnelling}.}


\subsubsection{RCV1/RCV2}\label{sec:comparable}

\noindent RCV1-v2 is a publicly available collection consisting of the
804,414 English news stories generated by Reuters from 20 Aug 1996 to
19 Aug 1997 \cite{Lewis:2004fk}.  RCV2 is instead a multilingual
collection, containing over 487,000 news stories
in one of thirteen languages other than English (Dutch, French,
German, Chinese, Japanese, Russian, Portuguese, Spanish,
LatinoAmerican Spanish, Italian, Danish, Norwegian, Swedish), and
generated by Reuters in the same timeframe. The documents of both
collections are classified according to the same hierarchically
organized set of 103 classes. The union of RCV1-v2 and RCV2 (hereafter
referred to as RCV1/RCV2) is a corpus \emph{comparable} at topic
level, as news stories are not direct translations of each other but
simply discuss the same or related events in different
languages. Since the corpus is not parallel, a training document for a
given language does not have, in general, a counterpart in the other
languages.

In our RCV1/RCV2 experiments we restrict our attention to the 9
languages (English, Italian, Spanish, French, German, Swedish, Danish,
Portuguese, and Dutch) for which stop word removal and lemmatization
are supported in
\texttt{NLTK}\footnote{\url{http://www.nltk.org/}}. In order to give
equal treatment to all these languages, from RCV1/RCV2 we randomly
select 1,000 training and 1,000 test news stories for each language
(with the sole exception of Dutch, for which only 1,794 documents are
available, and for which we thus select 1,000 documents for training
and 794 for test); this allows us to run our experiments in controlled
experimental conditions, i.e., to minimize the possibility that the
effects we observe across languages are due to different amounts of
training data for the different languages tested
upon.\footnote{\label{sec:caveat}The above selection protocol allows
us to \emph{minimize} the effects due to the amounts of training data
available for the different languages, but not to \emph{eliminate}
them. The reason is that different training examples may have
different number of classes associated to them, so one example that
has more of them contributes more training information than an example
that has fewer of them. This is a factor that is almost impossible to
eliminate from a multilabel dataset.}

Following this selection, we limit our consideration to the 73 classes
(out of 103) that end up having at least one positive training
example, in any of the 9 languages.  As a result, the average number
of classes per document is 3.21, ranging from a minimum of 1 to a
maximum of 13; the number of positive examples per class ranges from a
minimum of 1 to a maximum of 3,913. The average number of distinct
features (i.e., word lemmas) per language is 4,176, with a total of
26,977 distinct terms across all languages, of which 10,613 appear in
two or more languages.

Since the selection of 1,000 training and 1,000 test documents for
each language introduces a random factor, we repeat the entire process
10 times, each time with a different random selection; all the
RCV1/RCV2 results we report in this paper are thus averages across
these 10 random trials.


\subsubsection{JRC-Acquis}\label{sec:parallel}

\noindent JRC-Acquis (version 3.0) is a collection of parallel
legislative texts of European Union law written between the 1950s and
2006 \cite{Steinberger:2006rz}. JRC-Acquis is publicly available for
research purposes, and covers 22 official European languages. The
corpus is parallel and aligned at the sentence level, i.e., of each
document there are 22 language-specific versions which are
sentence-by-sentence translations of each other. The dataset is
labelled according to the EuroVoc thesaurus, which consists of a
hierarchy of more than 6,000 classes; for our experiments we select
the 300 most frequent ones.

We restrict our attention to the 11 languages (the same 9 languages of
RCV1/RCV2 plus Finnish and Hungarian) for which stop word removal and
lemmatization are supported in \texttt{NLTK} (we do not consider
Romanian due to incompatibilities found in the source files).

For inclusion in the training set we take all documents written in the
[1950,2005] interval and randomly select, for each of them, one of the
11 language-specific versions.
The rationale of this policy is to avoid the presence of
translation-equivalent content in the training set; this will enable
us to measure the contribution of training information coming from
different languages in a more realistic setting.

For the test set we instead take all documents written in 2006 and
retain all their 11 language-specific versions. The rationale behind
this policy is to allow a perfectly fair evaluation across languages,
since each of the 11 languages is thus evaluated on exactly the same
content. This process results in 12,687 training documents (between
1,112 and 1,198 documents per language) and 46,662 test documents
(exactly 4,242 documents per language). The average number of classes
per document is 3.31, ranging from a minimum of 1 to a maximum of 18;
the number of positive examples per class ranges from a minimum of 55
to a maximum of 1,155. There is an average of 9,909 distinct word
lemmas per language, a total of 81,458 distinct terms across all
languages, of which 27,550 appear in more than one language.

As in RCV1/RCV2, we repeat the process of selecting training data 10
times, each time with a different random selection (this means that,
in each of these 10 random trials, a different language-specific
version of the same document is selected); for JRC-Acquis too, all the
results we report in this paper are thus averages across these 10
random trials.


\subsection{Evaluation measures}\label{sec:evalmeasures}

\noindent As the evaluation measures for binary classification we use
both the ``classic'' $F_{1}$ and the more recently proposed $K$
\cite{Sebastiani:2015zl}.
These two functions are defined as
\begin{align}
  \label{eq:F1}
  F_{1} = & \ \left\{
            \begin{array}{cl}
              \dfrac{2TP}{2TP + FP + FN} & $if$ \ TP + FP + FN>0 \rule[-3ex]{0mm}{7ex} \\
              1 & $if$ \ TP=FP=FN=0 \\
            \end{array}
  \right.
  \\
  \label{eq:K} 
  K = & \ \left\{
        \begin{array}{cl}
          \dfrac{TP}{TP+FN}+\dfrac{\wasblue{TN}}{\wasblue{TN}+FP}-1 & $if$ \ TP+FN>0 \ $and$ \ TN+FP>0 \rule[-3ex]{0mm}{7ex} \\
          2\dfrac{\wasblue{TN}}{\wasblue{TN}+FP}-1 & $if$ \ TP+FN=0 \rule[-3ex]{0mm}{7ex} \\
          2\dfrac{TP}{TP+FN}-1 & $if$ \ TN+FP=0 
        \end{array}
                                 \right.
\end{align}
\noindent where $TP$, $FP$, $FN$, $TN$, represent the numbers of true
positives, false positives, false negatives, true negatives, generated
by a binary classifier. $F_{1}$ ranges between 0 (worst) and 1 (best);
$K$ ranges between -1 (worst) and 1 (best), with 0 corresponding to
the accuracy of the random classifier.

In order to turn $F_{1}$ and $K$ into measures for \emph{multilabel}
classification we compute their ``microaveraged'' versions (indicated
as $F_{1}^{\mu}$ and $K^{\mu}$) and their ``macroaveraged'' versions
(indicated as $F_{1}^{M}$ and $K^{M}$). $F_{1}^{\mu}$ and $K^{\mu}$
are obtained by (a) computing the class-specific values $TP_{j}$,
$FP_{j}$, $FN_{j}$, $TN_{j}$; (b) obtaining $TP$ as the sum of the
$TP_{j}$'s (same for $FP$, $FN$, $TN$), and then (c) applying
Equations \ref{eq:F1} and \ref{eq:K}. $F_{1}^{M}$ and $K^{M}$ are
obtained by first computing the class-specific values of $F_{1}$ and
$K$ and then averaging them across all $c_{j}\in\mathcal{C}$.

In all cases we also report the results of paired sample, two-tailed
t-tests at different confidence levels ($\alpha=0.05$ and
$\alpha=0.001$) in order to assess the statistical significance of the
differences in performance as measured by the averaged results.


\subsection{Representing text}\label{sec:representingtext}

\noindent We preprocess text by using the stop word removers and
lemmatizers available for all our languages within the
\texttt{scikit-learn}
framework\footnote{\url{http://scikit-learn.org/}}. As the weighting
criterion we use a version of the well-known $\mathit{tfidf}$ method,
expressed as
\begin{equation}
  \mathit{tfidf}(f,d)=\log\#(f,d)\times \log \frac{|Tr_{i}|}{|d'\in Tr_{i} : \#(f,d')>0|}
  \label{eq:tfidf}
\end{equation}
\noindent where $\#(f,d)$ is the raw number of occurrences of feature
$f$ in document $d$ and $\lambda_{i}$ is the language $d$ is written
in; weights are then normalized via cosine normalization, as
\begin{equation}
  w(f,d)=\frac{\mathit{tfidf}(f,d)}{\sqrt{\sum_{f'\in F_{i}} \mathit{tfidf}(f',d)^2}}
  \label{eq:tfidfnorm}
\end{equation}



\noindent Our feature spaces $F_{i}$ resulting from the different,
language-specific training sets $Tr_{i}$ are non-overlapping, since
(consistently with most multilingual text classification literature)
we do not make any attempt to detect matches between features across
different languages. Detecting such matches would be problematic,
since identical surface forms do not always translate to identical
meanings; e.g., while word \texttt{Madrid} as detected in a Spanish
text and word \texttt{Madrid} as detected in an Italian text may have
the same meaning, word \texttt{burro} as detected in a Spanish text
and word \texttt{burro} as detected in an Italian text typically do
not (\texttt{burro} means ``donkey'' in Spanish and ``butter'' in
Italian). The main reason why we do not attempt to detect such matches
is that neither funnelling (which uses different base classifiers for
the different languages) nor any of the baseline systems we use (see
Section \ref{sec:baselines}) would gain any advantage even from a
hypothetically perfect detection of such matches.


\subsection{Baselines}\label{sec:baselines}

\noindent We choose the following cross-lingual methods as the
baselines against which to compare our approach (see also Section
\ref{sec:relatedwork} for more detailed descriptions of these
methods):

\begin{itemize}

\item \textsc{Na\"ive:} This method consists in classifying each test
  document by a monolingual classifier trained on the corresponding
  language-specific portion of the training set; thus, there is no
  contribution from the training documents written in other
  languages. \textsc{Na\"ive}
  is usually considered a lower bound for any CLC effort.

\item \textsc{LRI:} \emph{Lightweight Random Indexing}
  \cite{Moreo:2016fk}, a CLC method that does not use any external
  resource.  In all experiments we set the dimensionality of the
  reduced space to 25,000.
  
\item \textsc{CLESA:} \emph{Cross-Lingual Explicit Semantic Analysis}
  \cite{Sorg:2012dn}. Unlike LRI and Funnelling, \textsc{CLESA} does
  require external resources, in the form of a comparable
  corpus of reference texts. In our experiments, consistently with the
  CLESA literature, as the reference texts we use 5,000 Wikipedia
  pages randomly chosen among the ones that (a) exist for all the
  languages in our datasets, and (b) contain 50 words or more in each
  of their language-specific versions. We use the Wikipedia Extractor
  tool\footnote{\url{http://medialab.di.unipi.it/wiki/Wikipedia_Extractor}}
  to obtain clean text versions of Wikipedia pages from a Wikipedia
  XML dump.  The tool filters out any other information or annotation
  present in Wikipedia pages, such as images, tables, references, and
  lists.

\item \wasblue{\textsc{KCCA:} \emph{Kernel Canonical Correlation
  Analysis} \cite{vinokourov2003inferring}.  We use the
  \texttt{Pyrcca} \cite{bilenko2016pyrcca} package to implement a
  cross-lingual classifier based on KCCA.  Since \texttt{Pyrcca} does
  not provide specialized data structures for storing sparse
  matrices\footnote{ \wasblue{\texttt{Pyrcca} is primarily optimized
  for working not on texts but on images. Still, it is the only
  available implementation we are aware of that allows to learn
  projections for more than two sets of variables.}
  }, 
  the amount of memory it requires in order to allocate all the
  language-specific
  views of the term co-occurrence matrices grows rapidly. In order to
  keep computation times within acceptable bounds, in our experiments
  we thus limit the number of comparable documents (for which we use
  Wikipedia articles, as for CLESA) to 2000 (and not 5000, as we do
  for \textsc{CLESA}).
  We set the number of components to 1000 and (after optimization via
  $k$-fold cross-validation) the regularization parameter to 1 for
  RCV1/RCV2 and to 10 for JRC-Acquis.  }

\item \textsc{DCI:} \emph{Distributional Correspondence Indexing}, as
  described in \cite{Moreo:2016fg}, and adapted to the cross-lingual
  setting by using the category labels (instead of a subset of terms)
  as the pivots.
  The dimensionality of the embedding space is thus set to the number
  of classes. In our experiments, as the distributional correspondence
  function (see \cite{Moreo:2016fg}) we adopt the linear one, since in
  preliminary experiments (not reported here for the sake of brevity)
  in which we used different such functions it proved the best one.
  
\item \textsc{MLE}: \emph{Multilingual Embeddings} derives document
  representations based on the multilingual word embeddings (of size
  300) released by \citet{Conneau:2018bv}. As proposed by the authors,
  documents are represented as an aggregation of the embeddings
  associated to the words they contain; since the word embeddings are
  aligned across languages, the documents end up being represented in
  the same vector space, irrespectively of the language they are
  written in. Given that we are representing documents (and not
  sentences as in \cite{Conneau:2018bv}), we weigh each embedding by
  its $\mathit{tfidf}$ score (instead of by its $\mathit{idf}$ score
  as suggested in \cite{Conneau:2018bv}), in order to better reflect
  the relevance of the term in the document (we have indeed verified
  $\mathit{tfidf}$ to perform better than simple $\mathit{idf}$ in
  preliminary experiments, which we do not discuss for the sake of
  brevity).
  
\item \wasblue{\textsc{MLE-LSTM}: Averaging embeddings causes a loss
  of word-order information. Modern NLP approaches attempt to capture
  such information by training Recurrent Neural Networks (RNNs) via
  ``backpropagation through time''. \textsc{MLE-LSTM} uses a Long
  Short-Term Memory (LSTM) cell \cite{hochreiter1997long} as the
  recurrent unit which, by processing sequences of embeddings,
  produces a document embedding that is then passed through a series
  of feed-forward connections with non-linear activations to finally
  derive a vector of probabilities for each class. The embeddings are
  initialized in \textsc{MLE-LSTM} with the multilingual embeddings
  released by \citet{Conneau:2018bv}, and are fine-tuned during
  training. We use 512 hidden units in the recurrent cell, and 2048
  units in the next-to-last feed-forward layer. The non-linear
  connection between layers is the ReLU (REctifier Linear Unit), and a
  0.5 dropout is applied to every layer and recurrent connections in
  order to prevent overfitting. We use the RMSprop optimizer
  \cite{graves2013generating} with default parameters to minimize the
  binary cross-entropy loss of the posterior probabilities with
  respect to the labels. We train the network through 200 epochs in
  RCV1/RCV2 and through 2000 epochs in JRC-Acquis, until convergence,
  with an early-stopping criterion that terminates the training after
  $p$ epochs show no improvement on the held-out validation set (a
  random sample containing 20\% of the training data); $p$ is the
  \emph{patience} parameter, that we set to 20 for RCV1/RCV2 and to
  200 for JRC-Acquis. Note that this is the only method among all the
  tested ones that accounts for word-order information.}

\item \textsc{UpperBound:} This is not a real (or realistic) baseline,
  but a system only meant to act, as the name implies, as an idealized
  upper bound that all CLC methods should strive to emulate (although
  its performance is hard to reach in practice). In
  \textsc{UpperBound} each non-English training example is replaced by
  its corresponding English version, a monolingual English classifier
  is trained, and all the English test documents are classified. We
  deploy \textsc{UpperBound} only for the JRC-Acquis dataset (where
  this gives rise to a training set of 12,687 English documents),
  since in RCV1/RCV2 the English versions of non-English training
  examples are not available.



\end{itemize}


\noindent Note that, despite the fact that ours is an ensemble
learning method, we do not include other such methods as
baselines. The reason is that other ensemble learning methods (such as
e.g., stacking, bagging, or boosting) inherently deal (as already
noted in Section \ref{sec:funnellingandstacking}) with ``homogeneous''
settings, i.e., scenarios in which all examples lie in the same
feature space. CLC is a ``heterogeneous'' setting, in which examples
written in different languages lie in different feature spaces, and
the above-mentioned methods are not equipped for dealing with these
scenarios. In fact, to the best of our knowledge, ours is the first
ensemble learning method in the literature that can deal with
heterogeneous settings.


\subsection{Learning algorithms}\label{sec:learning}

\noindent We have implemented our methods and all the baselines as
extensions of \texttt{scikit-learn}.

As the learning algorithm we use Support Vector Machines (SVMs), in
the implementation provided by \texttt{scikit-learn}. As customary in
multilabel classification, each 1st-tier multilabel classifier is
simply a set of independently trained binary classifiers, one for each
class $c\in\mathcal{C}$.

Note that, when training a \ftat\ classifier, when for a certain
$(\lambda_{i}, c_{j})$ pair there are no positive training examples,
we generate a trivial rejector, i.e., a classifier $h^{1}_{i}$ that
returns scores $h^{1}_{i}(d_{u},c_{j})=0$ (and, as a consequence,
posterior probabilities $\Pr(c_{j}|d_{u})=0$) for all test documents
$d_{u}$ written in language $\lambda_{i}$. In our datasets this can
indeed happen since, while we remove from both datasets the classes
that do not have any positive training examples, not all remaining
classes have positive training examples \emph{for every language}.

\wasblue{For the $k$-fold cross-validation needed in the \fkfcv\
method we use $k=10$.} We should also remark that, when training a
\fkfcv\ classifier, 
splitting the training set $Tr_{i}$ into $Tr_{i1},..., Tr_{ik}$ might
end up in placing all the positive training examples in the same
subset $Tr_{ix}$ (this always happens when there is a single positive
training example for $(\lambda_{i}, c_{j})$), which means that we
would be left with no positive training examples for training
classifier $h^{1}_{ix}$. In this case, instead of generating (as in
the \ftat\ case discussed above) a classifier $h^{1}_{ix}$ that works
as a trivial rejector, we train $h^{1}_{ix}$ via \ftat, i.e., by also
using the training examples in $Tr_{ix}$. In preliminary experiments
that we have carried out on a separate dataset, the use of this simple
heuristics has brought about substantial benefits; as a result we have
adopted it in all the experiments reported in this
paper.\footnote{\label{foot:stratif}\wasblue{One might wonder why, in
order to avoid the possibility that the union of $(k-1)$ folds
contains zero positive examples of a given class, when training
\fkfcv\ we do not use \emph{stratified} $k$-fold cross-validation
(which consists in choosing the $k$ folds in such a way that the class
prevalences in each fold are approximately equal to the class
prevalences in the entire training set). There are two reasons for
this. First, using stratification would not eradicate the problem,
because there are many pairs $(\lambda_{i},c_{j})$ for which there are
$\leq 1$ positive examples in the entire training set. Second,
stratification is convenient for binary or single-label
classification, but not for multilabel classification, where a
different split into $k$ folds must be set up for each different
class. For these reasons we opt for using the traditional
(non-stratified) variant.}}

We optimize the $C$ parameter, which controls the trade-off between
the training error and the margin of the SVM classifier, through a
5-fold cross-validation on the training set, via grid search on
$\{ 10^{-1},10^{0},\ldots , 10^{4}\}$; we do this optimization
individually for each method and for each run.  For the two funnelling
methods we perform this grid search only for the meta-classifier,
leaving $C$ to its default value of 1 for the base classifiers; the
main reason is that, especially in the case of \fkfcv\ (where an
expensive 10-fold cross validation is already performed in order to
generate the $\phi^{2}(d_{l})$ representations for the training
examples), the resulting computational cost would be severe.

Adhering to established practices in \wasblue{text classification} we
use two different kernels depending on the characteristics of the
feature space. For all classifiers operating in a high-dimensional and
sparse feature space (i.e., \textsc{UpperBound}, \textsc{LRI}, the
language-dependent classifiers of \textsc{Na\"ive}, plus the base
classifiers of the two funnelling methods) we use the linear kernel,
while we adopt the RBF kernel when the feature space is
low-dimensional and dense (i.e., for \textsc{CLESA},
\wasblue{\textsc{KCCA}}, \textsc{DCI}, \textsc{MLE}, and the
meta-classifier of the two funnelling methods).

For the two funnelling methods we use the probability calibration
algorithm implemented within \texttt{scikit-learn} and originally
proposed by Platt \cite{Platt:2000fk}, which consists of using, as the
mapping function $f$, a logistic function
\begin{equation}
  \label{eq:calibration}
  \Pr(c|d)=\dfrac{1}{1+e^{\alpha h(d,c)+\beta}}
\end{equation}
\noindent and choosing the parameters $\alpha$ and $\beta$ in such a
way as to minimize \wasblue{(via $k$-fold cross-validation)} the
negative log-likelihood of the \wasblue{training} data.






\section{Results}
\label{sec:results}


\subsection{Multilabel CLC experiments}
\label{sec:MLCLCresults}

\noindent Table \ref{tab:mlmc} shows our multilabel CLC results.  In
this table (and in all the tables of the next sections) each reported
value represents the average effectiveness across the 10 random
versions of each dataset (see Sections \ref{sec:comparable} and
\ref{sec:parallel}) and (with the exception of the \textsc{UpperBound}
values, which are computed on English test data only) across the
$|\mathcal{L}|$ languages in the dataset.  We report results for eight
combinations of (a) two datasets (RCV1/RCV2 and JRC-Acquis), (b) two
evaluation measures ($F_{1}$ and $K$), and (c) two different ways of
averaging the measure across the $|\mathcal{C}|$ classes of the
dataset (micro- and macro-averaging).

\begin{table}[tb]
  \centering \resizebox{\textwidth}{!} {
  \begin{tabular}{|c|l||c|c|c|c|c|c|c|c|c||c|}
    \hline
    \multicolumn{2}{|c||}{\mbox{}}
    & \begin{sideways}\textsc{Na\"ive}\end{sideways} 
    & \begin{sideways}LRI\end{sideways} 
    & \begin{sideways}CLESA\end{sideways} 
    & \begin{sideways}\wasblue{\textsc{KCCA}}\end{sideways}
    & \begin{sideways}DCI\end{sideways} 
    & \begin{sideways}MLE\end{sideways} 
    & \begin{sideways}\wasblue{MLE-LSTM}\end{sideways} 
    & \begin{sideways}\fkfcv\end{sideways} 
    & \begin{sideways}\ftat\end{sideways} 
    & \begin{sideways}\textsc{UpperBound}\phantom{0}\end{sideways} \\
    \hline
    \hline
    \multirow{2}{*}{$F_1^{\mu}$} 
    & RCV1/RCV2 & .776 $\pm$ .052 &  .771  $\pm$ .050 & .714  $\pm$ .061 & \wasblue{.616 $\pm$ .065}	 & .770  $\pm$ .052 & .696\psddag  $\pm$ .060 & \wasblue{.574 $\pm$ .113} & .801\sdag  $\pm$ .044 &\cellcolor[gray]{.7}\textbf{.802} $\pm$ .041 & -- \\ 
    \cline{2-12}
    & JRC-Acquis & .559  $\pm$ .012 & \cellcolor[gray]{.7}\textbf{.594}  $\pm$ .016 & .557  $\pm$ .024 & \wasblue{.357 $\pm$ .023} & .510  $\pm$ .014 &  .478\psddag  $\pm$ .061 & \wasblue{.378 $\pm$ .041} & .581\psdag  $\pm$ .010 & .587  $\pm$ .009 & .707 \\
    \hline
    \hline
    \multirow{2}{*}{$F_1^{M}$}
    & RCV1/RCV2 & .467  $\pm$ .083 & .490  $\pm$ .077	& .471  $\pm$ .074 & \wasblue{.385 $\pm$ .079} & .485  $\pm$ .070 & .453\psddag  $\pm$ .060 & \wasblue{.302 $\pm$ .115} & .512\psdag  $\pm$ .067 & \cellcolor[gray]{.7}\textbf{.534}  $\pm$ .066 & -- \\ 
    \cline{2-12}
    & JRC-Acquis & .340  $\pm$ .017 & \cellcolor[gray]{.7}\textbf{.411}  $\pm$ .027 & .379  $\pm$ .034 & \wasblue{.206 $\pm$ .018} & .317  $\pm$ .012 &  .300\psddag  $\pm$ .065 & \wasblue{.182 $\pm$ .030} & .356\psdag\  $\pm$ .013 & .399  $\pm$ .013 & .599   \\

    \hline
    \hline
    \multirow{2}{*}{$K^{\mu}$} 
    & RCV1/RCV2 & .690  $\pm$ .074 & .696  $\pm$ .069 & .659  $\pm$ .075 & \wasblue{.550 $\pm$ .073} & .696  $\pm$ .065 & .644\psddag\  $\pm$ .070 & \wasblue{.515 $\pm$ .127} & .731\psdag\  $\pm$ .058 &  \cellcolor[gray]{.7}\textbf{.760} $\pm$ .052 & -- \\ 
    \cline{2-12}
    & JRC-Acquis  & .429  $\pm$ .015 & .476  $\pm$ .020 & .453  $\pm$ .029 & \wasblue{.244 $\pm$ .022}	& .382  $\pm$ .016 & .429\psddag\  $\pm$ .050 & \wasblue{.292 $\pm$ .046} & .457\psdag\  $\pm$ .012 &  \cellcolor[gray]{.7}\textbf{.490}  $\pm$ .013 & .632 \\
    \hline
    \hline
    \multirow{2}{*}{$K^{M}$}
    & RCV1/RCV2 & .417  $\pm$ .090 & .440  $\pm$ .086 & .434  $\pm$ .080 & \wasblue{.358 $\pm$ .088} & .456  $\pm$ .082 & .466\psddag\  $\pm$ .073 & \wasblue{.280 $\pm$ .118} & .482\psdag\  $\pm$ .075 & \cellcolor[gray]{.7}\textbf{.506}  $\pm$ .073 & -- \\ 
    \cline{2-12}
    & JRC-Acquis & .288  $\pm$ .016 & .348  $\pm$ .025 & .330  $\pm$ .034 & \wasblue{.176 $\pm$ .017} & .274  $\pm$ .013 & .349\sddag\  $\pm$ .047 & \wasblue{.170 $\pm$ .032} & .328\psdag\  $\pm$ .013 & \cellcolor[gray]{.7}\textbf{.365}  $\pm$ .014 & .547 \\
    \hline
  \end{tabular}
  }
  \caption{Multilabel CLC results; \wasblue{each cell indicates the
  value for the effectiveness measure and the standard deviation
  across the 10 runs}. A greyed-out cell with a value in
  \textbf{boldface} indicates the best method (with the exclusion of
  \textsc{UpperBound}). Superscripts $\dag$ and $\dag\dag$ denote the
  method (if any) whose score is not statistically significantly
  different from the best one at $\alpha=0.05$ ($\dag$) or at
  $\alpha=0.001$ ($\dag\dag$).  }
  \label{tab:mlmc}
\end{table}

The results clearly indicate that our two funnelling methods perform
very well. In particular, \ftat\ is the best performer in 6 out of 8
combinations of dataset, evaluation measure, averaging method, always
outperforming all competitors in terms of the $K$ measure and on the
RCV1/RCV2 dataset. The only exception to this superiority is recorded
for $F_{1}^{\mu}$ and $F_{1}^{M}$ on the JRC-Acquis dataset, where
\textsc{LRI} is the best method; note, however, that in these cases
\textsc{LRI} outperforms \ftat\ only by a moderate margin, while in
the previously discussed 6 cases the superiority of \ftat\ is more
marked. In 8 out of 8 cases \ftat\ outperforms \textsc{Na\"ive},
\textsc{CLESA}, \wasblue{\textsc{KCCA}}, \textsc{DCI}, \textsc{MLE},
\wasblue{and \textsc{MLE-LSTM}}, almost always by a very wide margin.

The experiments also indicate that the simpler \ftat\ is consistently
better than \fkfcv, with the former outperforming the latter in all 8
cases. Together with the fact that \ftat\ is markedly cheaper (by a
factor of $(k+1)$) to train than \fkfcv, this makes \ftat\ our method
of choice.

\begin{figure}[tb]
  \centering
  \includegraphics[width=\textwidth]{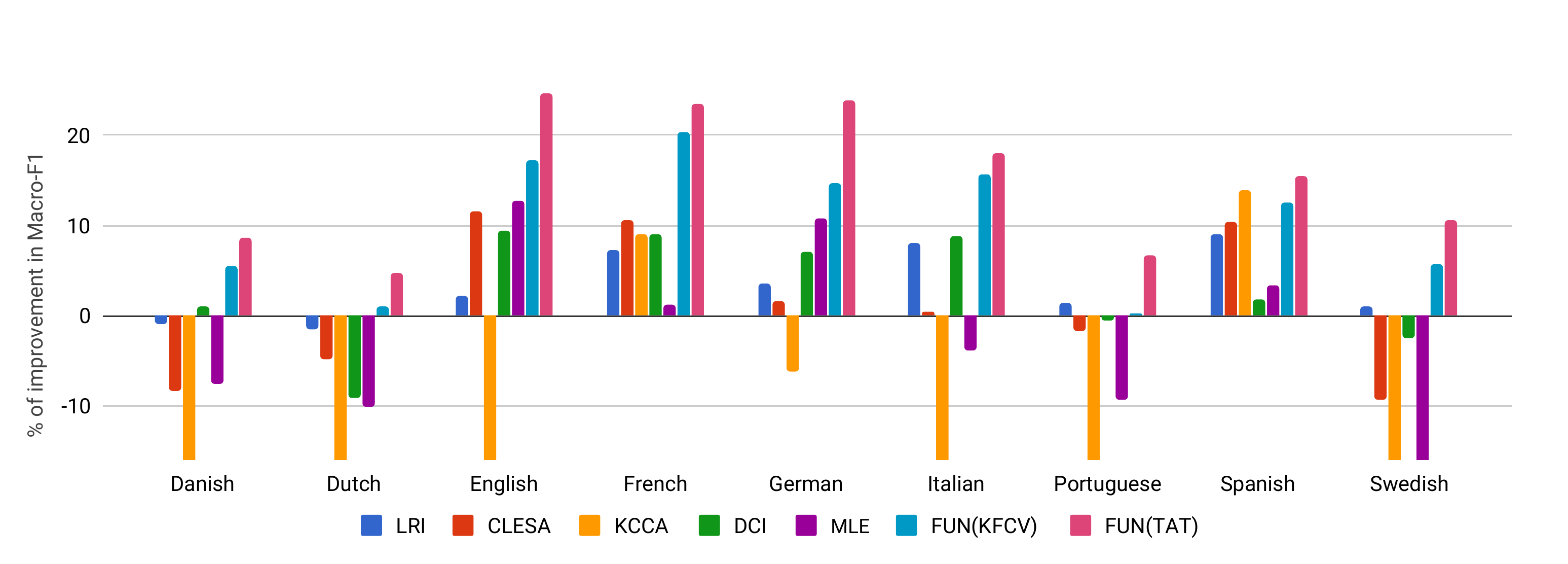}
  \caption{Per-language percentage improvement in $F_1^{M}$ with
  respect to each \textsc{Na\"ive} monolingual classifier in
  RCV1/RCV2. \wasblue{Some methods (notably: KCCA and MLE) sometimes
  exhibit deteriorations so large that they would be difficult to
  display in full; in these cases, bars are truncated at approximately
  -15\% deterioration.} }
  \label{fig:perlang}
\end{figure}

As already mentioned, the results displayed in Table \ref{tab:mlmc}
are averages across the $|\mathcal{L}|$ languages in the
dataset. Analysing the results in a finer-grained way (that is, on a
language-by-language basis) shows a further interesting fact: \ftat\
and \fkfcv\ are the only systems that outperform the \textsc{Na\"ive}
baseline \emph{in every case}, i.e., for each language, dataset,
evaluation measure, and averaging method (micro- or macro-). An
example of this fact is shown in Figure \ref{fig:perlang},
which displays the percentage improvement (in terms of $F_{1}^{M}$)
obtained by the various methods with respect to the \textsc{Na\"ive}
baseline for the various languages on the RCV1/RCV2 dataset. The
figure shows that CLESA, DCI, \wasblue{KCCA,} MLE, and even LRI
(according to Table \ref{tab:mlmc}, the best competitor of funnelling
methods), perform worse than \textsc{Na\"ive} for some languages,
while both \ftat\ and \fkfcv\ outperform \textsc{Na\"ive} for all
languages. \wasblue{MLE-LSTM is not included in this plot since it
always underperforms \textsc{Na\"ive} by such a large margin that
including it in the plot would substantially hinder the visualization
of the other results.}  \ftat\ thus proves not only the best method of
the lot, but also the most stable.

\wasblue{That \textsc{KCCA} underperforms \textsc{CLESA} on most
languages might be explained by the reduction in the number of
Wikipedia articles that \textsc{KCCA} has observed (for the reasons
discussed in Section \ref{sec:baselines}) during training with respect
to \textsc{CLESA}.}  \wasblue{Concerning \textsc{MLE}, instead, it is
immediate to observe that it does not perform well, in many cases
underperforming the \textsc{Na\"ive} baseline.  A possible reason for
this might reside in the fact that \textsc{MLE} was originally devised
for (and showed good performance on) \emph{sentence} classification;
it is easy to conjecture that, when the units of classification are
(as here) linguistic objects much longer than sentences, a method that
just computes averages across word embeddings might introduce more
noise than information.}  \wasblue{Regarding \textsc{MLE-LSTM}, we
conjecture that its very bad performance might be explained by two
facts.  First, many words from different languages are not covered in
the pre-trained multilingual embeddings; those words, that are instead
initialized with zero-embeddings\footnote{\wasblue{We have tested
other approaches including random initialization, or replacing them
with a language-specific \emph{unknown} token. None of them
effectively help to improve the results.}}, might affect negatively
the entire optimization procedure.
Second, it is very likely that 
the training set for each language 
is too small for a deep model to find meaningful cross-lingual
patterns, thus making the classifier suffer from noisy information.}

Incidentally, Figure \ref{fig:perlang} shows that the language on
which \ftat\ obtains the highest $F_{1}^{M}$ improvement on RCV1/RCV2
with respect to the \textsc{Na\"ive} baseline, is English (in Table
\ref{tab:languagebenefit} we show this fact to hold in RCV1/RCV2
irrespectively of evaluation measure and averaging method).  This
shows that CLC techniques, and funnelling techniques in particular,
can also benefit languages that are often considered ``easy'' (since
they have historically received more attention than others from the
research community), and for which obtaining improvements is thus
considered harder.

An interesting observation we can make by observing Table
\ref{tab:mlmc} is that (a) \textsc{UpperBound} always works better
than \ftat\ and \fkfcv, and (b) \ftat\ and \fkfcv\ always work better
than \textsc{Na\"ive}. Fact (a) indicates that the standard ``bag of
words'', content-based representations which \textsc{UpperBound} uses
work better than the representations based on posterior probabilities
that \ftat\ and \fkfcv\ use, because \textsc{UpperBound}, \ftat\ and
\fkfcv\ use exactly the same training examples (i.e., the examples in
$\bigcup_{i=1}^{|\mathcal{L}|}Tr_{i}\}$), although represented
differently. However, fact (b) shows that the inferior quality of the
latter representations is more than compensated by the availability of
many additional training examples, since \textsc{Na\"ive} uses a small
subset ($|\mathcal{L}|$ times smaller) of the set of training examples
that \ftat\ and \fkfcv\ use.



  




\subsection{Multilabel monolingual and binary cross-lingual
experiments}\label{sec:monolingualresults}

\noindent As discussed in Section \ref{sec:funnellingandstacking}, we
conjecture that the good performance obtained by funnelling in the
multilabel CLC experiments partly derives from the fact that the
stochastic dependencies between the classes are brought to bear, and
partly derives from the ability of funnelling to leverage training
data written in language $\lambda^{s}$ for classifying the data
written in language $\lambda^{t}$.  In order to verify if both factors
indeed contribute to multilabel CLC, we run multilabel monolingual
experiments and binary cross-lingual experiments.

%

In our multilabel monolingual experiments a funnelling system tackles
a single language $\lambda_{i}$, i.e., there is just one 1st-tier
multilabel classifier $h_{i}^{1}$ and the meta-classifier is trained
only from the documents in $Tr_{i}$ (instead of all the documents in
$\bigcup_{i=1}^{|\mathcal{L}|}Tr_{i}$, as was the case in Section
\ref{sec:MLCLCresults}).  (Note that, in this particular setting,
stacking and funnelling coincide, as there is no heterogeneity in the
data.)  With such a setup, any improvement with respect to the
\textsc{Na\"ive} baseline can only be due to the fact that funnelling
brings to bear the stochastic dependencies between the classes. We run
multilabel monolingual experiments independently for all the
$|\mathcal{L}|$ languages in the dataset. The results (reported as
averages across these $|\mathcal{L}|$ languages) are displayed in
Column B of Table \ref{tab:monolingualbinary}.

In our binary cross-lingual experiments, instead, a funnelling system
tackles a single class, i.e., the $\phi^{2}(d_{u})$ vectors fed to the
meta-classifier only consist of one posterior probability (instead of
$|\mathcal{C}|$ posterior probabilities, as was the case in Section
\ref{sec:MLCLCresults}), so that any improvement with respect to the
\textsc{Na\"ive} baseline can only be due to the ability of funnelling
to leverage training data written in language $\lambda^{s}$ for
classifying the data written in language $\lambda^{t}$. We run binary
cross-lingual experiments independently for all the $|\mathcal{C}|$
classes in the dataset. The results are displayed in Column C of Table
\ref{tab:monolingualbinary}.

Note that in these experiments (a) we do not run LRI, CLESA, DCI, and
MLE, since our only goal here is to assess where the improvements of
funnelling with respect to the \textsc{Na\"ive} baseline come from;
(b) we only run \ftat\, since its superiority with respect to \fkfcv\
has already been ascertained in a fairly conclusive way in Section
\ref{sec:MLCLCresults};
(c) in Table \ref{tab:monolingualbinary} (as, for that matter, in
all other tables in this paper) the results reported in the 4 columns
for the same row are all comparable with each other, since the
training set and the test set are the same in all 4 cases.

The results of Table \ref{tab:monolingualbinary} suggest the following
observations:

\begin{enumerate}

\item \label{item:multilabel} Using \ftat\ in order to bring to bear
  the stochastic dependencies between different classes is useful, as
  witnessed by the fact that the figures for the multilabel
  monolingual setup are always higher than the corresponding figures
  for the \textsc{Na\"ive} baseline.

\item \label{item:cross-lingual} Using \ftat\ in order to leverage
  training data written in one language for classifying the data
  written in other languages, is also useful, as witnessed by the fact
  that the figures for the binary cross-lingual setup are always
  higher than the corresponding figures for the \textsc{Na\"ive}
  baseline.

\item The two observations above are confirmed by the fact that the
  figures for the multilabel cross-lingual setup are (almost always)
  higher than the figures for both the multilabel monolingual and the
  binary cross-lingual setups. In other words, \emph{both} factors
  contribute to the fact that \ftat\ in the multilabel cross-lingual
  setup improves on the \textsc{Na\"ive} baseline.

\item While both factors do contribute, it is also clear that the
  bigger contribution comes not from \wasblue{the stochastic
  dependencies between different classes}, but from \wasblue{the
  training data in other languages},
  as witnessed by the fact that the figures for the multilabel
  cross-lingual setup are much closer to the binary cross-lingual ones
  than to the multilabel monolingual ones.

\end{enumerate}


\begin{table}[tb]
  \centering
  \begin{tabular}{|c|l||c|c|c|c|c|}
    \hline
    \multicolumn{2}{|c||}{\mbox{}} & A & B & C & D \\
    \hline
    \multicolumn{2}{|c||}{\mbox{}} & \textsc{Na\"ive} & \ftat\ & \ftat\ & \ftat\ \\ 
    \multicolumn{2}{|c||}{\mbox{}} & Binary & MultiLab & Binary & MultiLab \\
    \multicolumn{2}{|c||}{\mbox{}} & MonoLin & MonoLin & CrossLin & CrossLin \\
    \hline
    \hline
    \multirow{2}{*}{$F_1^{\mu}$} 
                                   & RCV1/RCV2 & .776  $\pm$ .052 & .800\sddag\ $\pm$ .002 &  .801\sddag\  $\pm$ .002 & \cellcolor[gray]{.7}  \textbf{.802}\psddag\  $\pm$ .041 \\    \cline{2-6}
                                   & JRC-Acquis & .559  $\pm$ .012 & .577\psddag\ $\pm$ .002  & \cellcolor[gray]{.7}\textbf{.589}\psddag\  $\pm$ .002 & .587\sddag\  $\pm$ .009 \\    \hline    \hline
    \multirow{2}{*}{$F_1^{M}$}
                                   & RCV1/RCV2 & .467  $\pm$ .083 & .526\psddag\ $\pm$ .013 & .532\sdpsdag\  $\pm$ .014 & \cellcolor[gray]{.7}\textbf{.534}\psddag\  $\pm$ .066\\    \cline{2-6}
                                   & JRC-Acquis & .340  $\pm$ .017 & .369\psddag\ $\pm$ .002 & .395\sddag\  $\pm$ .003 & \cellcolor[gray]{.7}\textbf{.399}\psddag\  $\pm$ .013\\    \hline    \hline
    \multirow{2}{*}{$K^{\mu}$} 
                                   & RCV1/RCV2 & .690  $\pm$ .074 & .747\psddag\ $\pm$ .003 & .757\psddag\ $\pm$ .004 & \cellcolor[gray]{.7}\textbf{.760}\psddag\  $\pm$ .052\\    \cline{2-6}
                                   & JRC-Acquis & .429  $\pm$ .015 & .454 \psddag\ $\pm$ .002 & .487\sddag\ $\pm$ .002 & \cellcolor[gray]{.7}\textbf{.490}\psddag\  $\pm$ .013\\    \hline    \hline
    \multirow{2}{*}{$K^{M}$}
                                   & RCV1/RCV2 & .417  $\pm$ .090 & .492\psddag\ $\pm$ .013 & .505\sdpsdag\ $\pm$ .014 & \cellcolor[gray]{.7}\textbf{.506}\psddag\  $\pm$ .073\\    \cline{2-6}
                                   & JRC-Acquis & .288  $\pm$ .016 & .325\psddag\ $\pm$ .003 & .359\psddag\ $\pm$ .003 & \cellcolor[gray]{.7}\textbf{.365}\psddag\  $\pm$ .014\\    \hline
  \end{tabular}
  \caption{\ftat\ results for 
  multilabel monolingual classification (Column B) and binary
  cross-lingual classification (Column C). The results in Columns A
  and D are from Table \ref{tab:mlmc}, and are reported here only for
  ease of comparison.  The notational conventions are the same as in
  Table \ref{tab:mlmc}.}
  \label{tab:monolingualbinary}
\end{table}


\subsection{Learning curves for the under-resourced
languages}\label{sec:lessresourced}

\noindent
As we have mentioned in the introduction, CLC techniques are
especially useful when we need to perform text classification for
under-resourced languages, i.e., languages for which only a small
number of training documents are available. In this section we provide
the results of experiments aimed at showing how funnelling performs in
such situations. We simulate these scenarios by testing, on the
$\lambda_{i}$ test data, a \ftat\ system trained on all the training
data for the languages in $\mathcal{L}/\{\lambda_{i}\}$ and on
variable fractions of the training data for $\lambda_{i}$, which thus
plays (especially when these fractions are small) the role of the
under-resourced language. When this fraction is 0\% \wasblue{of the
total}, this corresponds to the \wasblue{zero-shot} setting; when it is
100\% \wasblue{of the total}, this corresponds to the setup we have
studied in Section \ref{sec:MLCLCresults}. In our experiments we
generate these fractions by randomly removing increasing amounts of
data from the training set, so that the training sets for the smaller
fractions are proper subsets of those for the larger fractions. Like
for all other experiments in this paper, the results we report are
averages across the 10 random trials discussed at the end of Sections
\ref{sec:comparable} and \ref{sec:parallel}.

\begin{figure}[tb]
  \centering
  \includegraphics[width=1.0\textwidth]{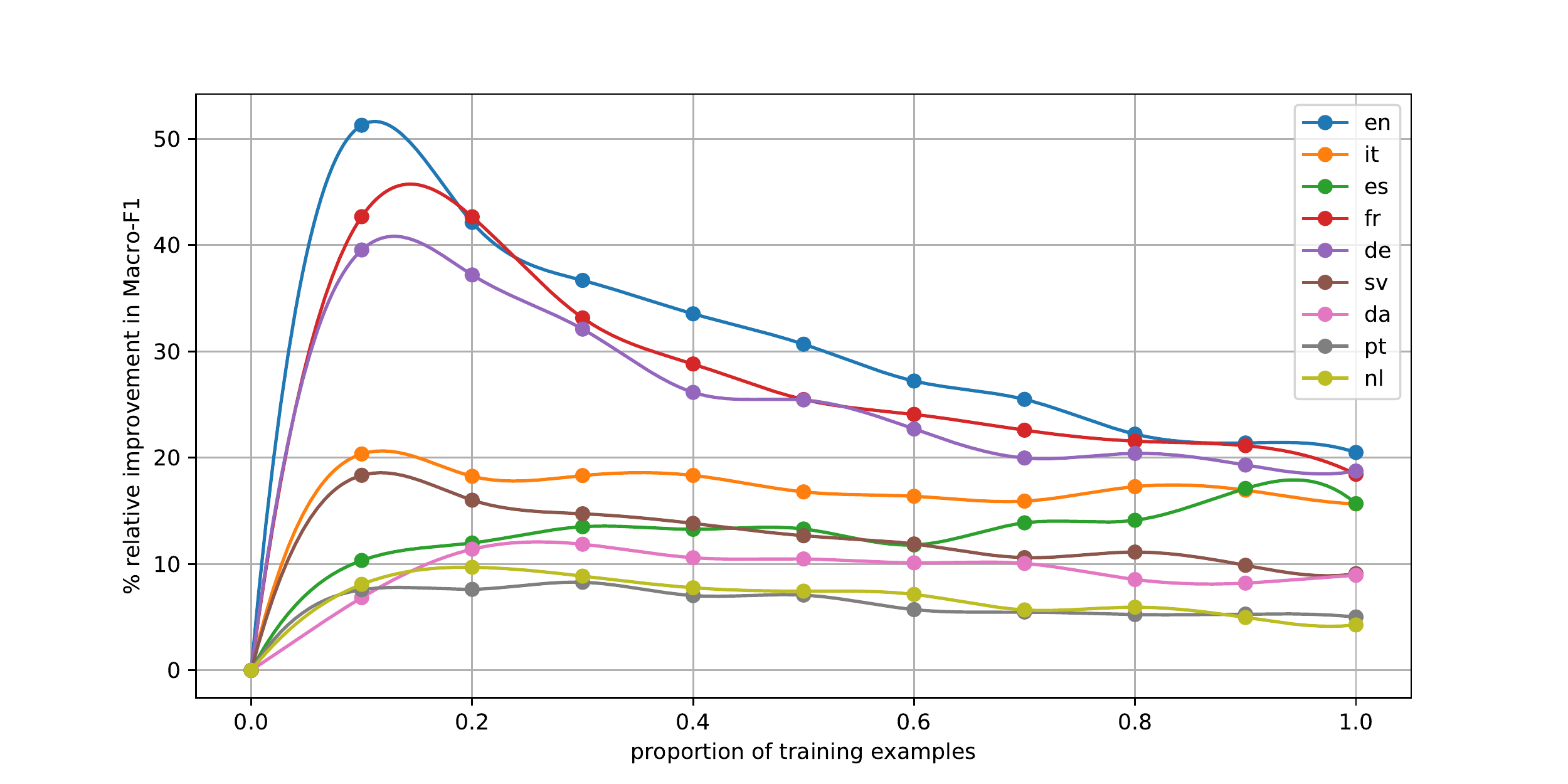}
  \caption{Relative improvement in terms of $F_{1}^{M}$ on the
  RCV1/RCV2 dataset obtained by using \ftat\ with respect to using
  \textsc{Na\"ive}. Values on the $x$ axis are the fractions of
  $Tr_{i}$ that are used for training.}
  \label{fig:relimp}
\end{figure}

Figure \ref{fig:relimp} shows,
for the RCV1/RCV2 dataset and the $F_{1}^{M}$ measure (the cases of
JRC-Acquis and/or the other measures show similar patterns), the
improvements which are obtained on the test sets of the individual
languages $\lambda_{i}$ as a function of the fraction of the training
data $Tr_{i}$ used. There are three main observations that we can
make: (a) for each language $\lambda_{i}$ and each fraction of
training data used, the variation in accuracy is always positive,
i.e., there is always an improvement in accuracy (and never a
deterioration) as a result of using funnelling; (b) some languages
benefit more than others (in our case, English, French, and German
stand out in this respect); (c) the improvements are more marked
when small fractions of $\lambda_{i}$ training data are used. Fact
(b) will be the subject of detailed study in Section
\ref{sec:languages}. As for Fact (c), this is intuitive after all,
since it is when the accuracy of a monolingual classifier is low (as
it presumably is when it has been trained from few labelled data) that
the margins of improvement resulting from the contributions of other
languages are high.


\subsection{Which languages contribute/benefit
most?}\label{sec:languages}


\noindent
In this section we present ``ablation'' experiments in which we
attempt to understand (a) which languages contribute most, and (b)
which languages benefit most, in terms of the classification
effectiveness that can be obtained via \ftat\ in multilabel CLC. In
order to do this, for each pair of languages
$\lambda^{s},\lambda^{t} \in \mathcal{L}$ we classify the
$\lambda^{t}$ test data via (a) a \ftat\ system trained on
$\mathcal{L}/\{\lambda^{s}\}$ training data, and \wasblue{(b)} a
\ftat\ system trained on $\mathcal{L}$ training data. The improvement
$i(\lambda^{s},\lambda^{t})$ observed in switching from (a) to (b) is
a measure of the contribution that $\lambda^{s}$ training data offer
to classifying $\lambda^{t}$ data, or (said another way) of the
benefit that the classification of $\lambda^{t}$ data obtains from the
presence of $\lambda^{s}$ training data. Similarly to what we have
done in Section \ref{sec:lessresourced}, in all these experiments we
adopt an ``under-resourced language'' setting and use only 10\% of the
$\lambda^{t}$ training examples. Note that the notion of ``improvement
in effectiveness'' mentioned above depends on which measure of
effectiveness (among the four we have employed in this paper) we use
as reference.

Displaying all the $|\mathcal{L}|\times|\mathcal{L}|$ individual
$i(\lambda^{s},\lambda^{t})$ results would probably not allow
significant insights to be obtained. However, in our multilabel CLC
context they can be aggregated so as to measure

\begin{enumerate}

\item which languages contribute most to the classification of data in
  other languages; we compute the contribution $\alpha(\lambda^{s})$
  of language $\lambda^{s}$ as
  the average value of $i(\lambda^{s},\lambda^{t})$ across all
  $\lambda^{t}\in \mathcal{L}/\{\lambda^{s}\}$;

\item which languages benefit most from the presence of training data
  in other languages; we compute the benefit $\beta(\lambda^{t})$ that
  language $\lambda^{t}$ obtains as the average value of
  $i(\lambda^{s},\lambda^{t})$ across all
  $\lambda^{s}\in \mathcal{L}/\{\lambda^{t}\}$.

\end{enumerate}
\noindent These results are reported in Tables
\ref{tab:languagecontribution} and \ref{tab:languagebenefit}. Rather
than commenting on the individual cases, one interesting question we
may ask ourselves is: \wasblue{what are the factors that make} a
language contribute more, or benefit more, within a funnelling system
for CLC?  Are there interesting correlations between these
contributions / benefits and other measurable characteristics of the
individual languages? Note that all languages have the same number of
training examples (and they also have the same number of test
examples), both in \textsc{RCV1/RCV2} and \textsc{JRC-Acquis}, so
(even considering what we say in Footnote \ref{sec:caveat}) language
frequency is unlikely to be a factor in our experiments.

\wasblue{A first conjecture we test} is if the contribution
$\alpha(\lambda^{s})$ is positively correlated with the accuracy of
the \textsc{Na\"ive} classifier for language $\lambda^{s}$ as computed
on $\lambda^{s}$ test data (we here denote this accuracy as
$F_{1}^{M}(\textsc{Na\"ive}(\lambda^{s}))$).\footnote{As in Section
\ref{sec:lessresourced}, as the measure of accuracy we here employ
$F_{1}^{M}$ in computing both $\alpha(\lambda^{s})$ and the accuracy
of the \textsc{Na\"ive} classifier for language $\lambda^{s}$; the
other measures used in this paper display similar results.}
This conjecture would seem sensible, since we would expect the
contribution of a language to be high when its language-specific
training data are high-quality (which is witnessed by the fact that a
classifier trained on them is capable of delivering high accuracy). We
measure correlation via the \emph{Pearson Correlation Coefficient}
(PCC), noted as $\rho(X,Y)$; its values range on [-1,+1], with -1
indicating perfect negative correlation, +1 indicating perfect
positive correlation, and 0 indicating total lack of correlation. The
above conjecture proves essentially correct, since the resulting value
of PCC is
$\rho(\alpha(\lambda^{s}),F_{1}^{M}(\textsc{Na\"ive}(\lambda^{s})))=0.788$
(with a p-value of 0.011), which indicates high
correlation.\footnote{For PCC, the p-value indicates the probability
that two random variables that have no correlation generate a sample
characterized by a value of PCC at least as extreme as the one of the
present sample.}

\begin{table*}[tb]
  \centering \resizebox{\textwidth}{!} {
  \begin{tabular}{|c|l||r|r|r|r|r|r|r|r|r|r|r|}
    \hline
    \multicolumn{2}{|c||}{\mbox{}} 
                                     & \multicolumn{1}{c|}{EN}  
                                     & \multicolumn{1}{c|}{IT}  
                                     & \multicolumn{1}{c|}{ES}  
                                     & \multicolumn{1}{c|}{FR}  
                                     & \multicolumn{1}{c|}{DE}  
                                     & \multicolumn{1}{c|}{SV}  
                                     & \multicolumn{1}{c|}{DA}  
                                     & \multicolumn{1}{c|}{PT}  
                                     & \multicolumn{1}{c|}{NL}  
                                     & \multicolumn{1}{c|}{FI}  
                                     & \multicolumn{1}{c|}{HU}  \\
    \hline
    \hline
    \multirow{2}{*}{$F_1^{\mu}$} 
                                     & RCV1/RCV2 & +0.08\% & +0.68\% & +0.34\% & +0.49\% & +0.03\% & \cellcolor[gray]{.7}\textbf{+2.25}\% & +0.06\% & +0.41\% & +0.18\%  & -- & -- \\ \cline{2-13}
                                     & JRC-Acquis & -0.11\% & +2.85\% & -0.20\% & +0.67\% & +0.01\% & -0.56\% & -0.12\% & +2.67\% & +\cellcolor[gray]{.7}\textbf{3.35}\% & +0.03\% & +1.84\% \\ \hline
    \multirow{2}{*}{$F_1^{M}$}
                                     & RCV1/RCV2 & -0.05\% & +0.36\% & -0.00\% & +0.11\% & +0.04\% & +0.75\% & +0.17\% & \cellcolor[gray]{.7}\textbf{+1.19}\% & +0.82\%  & -- & -- \\ \cline{2-13}
                                     & JRC-Acquis & -0.64\% & +5.98\% & -0.95\% & +0.83\% & -0.45\% & -10.23\% & -0.37\% & +3.61\% & +\cellcolor[gray]{.7}\textbf{6.23}\% & -0.60\% & +3.76\% \\ \hline
    \multirow{2}{*}{$K^{\mu}$} 
                                     & RCV1/RCV2 & +0.70\% & +1.52\% & +0.99\% & +0.41\% & +1.12\% & \cellcolor[gray]{.7}\textbf{+7.71}\% & +0.74\% & +2.91\% & +1.65\%  & -- & -- \\ \cline{2-13}
                                     & JRC-Acquis & +0.80\% & +7.85\% & +1.07\% & +3.63\% & +0.63\% & +2.67\% & +0.16\% & +7.78\% & +\cellcolor[gray]{.7}\textbf{8.83}\% & +1.85\% & +5.90\% \\ \hline
    \multirow{2}{*}{$K^{M}$}
                                     & RCV1/RCV2 & +0.39\% & +1.03\% & +0.45\% & +0.39\% & +0.60\% & \cellcolor[gray]{.7}\textbf{+3.55}\% & +0.46\% & +2.81\% & +2.07\%  & -- & -- \\ \cline{2-13}
                                     & JRC-Acquis & +0.99\% & +10.97\% & +1.37\% & +4.74\% & +0.66\% & -1.23\% & +0.15\% & +9.20\% & +\cellcolor[gray]{.7}\textbf{11.65}\% & +2.30\% & +8.36\% \\ \hline
  \end{tabular}
  }
  \caption{Average contribution (across languages
  $\lambda^{t}\in \mathcal{L}/\{\lambda^{s}\}$) provided by
  $\lambda^{s}$ training data to classifying $\lambda^{t}$ test data
  via \ftat. A greyed-out cell with a value in \textbf{boldface}
  indicates the language that has contributed most.}
  \label{tab:languagecontribution}
\end{table*}

\begin{table*}[tb]
  \centering \resizebox{\textwidth}{!} {
  \begin{tabular}{|c|l||r|r|r|r|r|r|r|r|r|r|r|}
    \hline
    \multicolumn{2}{|c||}{\mbox{}}                                                                   & \multicolumn{1}{c|}{EN}  
                                                                                                     & \multicolumn{1}{c|}{IT}  
                                                                                                     & \multicolumn{1}{c|}{ES}  
                                                                                                     & \multicolumn{1}{c|}{FR}  
                                                                                                     & \multicolumn{1}{c|}{DE}  
                                                                                                     & \multicolumn{1}{c|}{SV}  
                                                                                                     & \multicolumn{1}{c|}{DA}  
                                                                                                     & \multicolumn{1}{c|}{PT}  
                                                                                                     & \multicolumn{1}{c|}{NL}  
                                                                                                     & \multicolumn{1}{c|}{FI}  
                                                                                                     & \multicolumn{1}{c|}{HU}  
    \\
    \hline
    \hline
    \multirow{2}{*}{$F_1^{\mu}$} 
                                     & RCV1/RCV2 & \cellcolor[gray]{.7}\textbf{+1.70}\% & +0.76\% & +0.39\% & +0.95\% & +0.72\% & +0.08\% & -0.26\% & +0.20\% & -0.01\%  & -- & -- \\ \cline{2-13}
                                     & JRC-Acquis & +0.17\% & +1.22\% & +1.59\% & +1.31\% & +2.20\% & -0.45\% & +1.16\% & +0.53\% & +0.61\% & +\cellcolor[gray]{.7}\textbf{2.27}\% & -0.17\% \\ \hline
    \multirow{2}{*}{$F_1^{M}$}
                                     & RCV1/RCV2 & \cellcolor[gray]{.7}\textbf{+2.98}\% & +0.27\% & +0.13\% & -0.12\% & +0.56\% & -0.06\% & -0.62\% & +0.21\% & +0.04\%  & -- & -- \\ \cline{2-13}
                                     & JRC-Acquis & +1.13\% & +\cellcolor[gray]{.7}\textbf{2.12}\% & +1.08\% & +1.22\% & +1.55\% & -2.16\% & +1.05\% & +0.57\% & +1.17\% & +0.75\% & -1.31\% \\ \hline
    \multirow{2}{*}{$K^{\mu}$} 
                                     & RCV1/RCV2 & \cellcolor[gray]{.7}\textbf{+3.33}\% & +2.73\% & +1.81\% & +2.40\% & +2.41\% & +1.26\% & +0.12\% & +2.60\% & +1.10\%  & -- & -- \\ \cline{2-13}
                                     & JRC-Acquis & +3.06\% & +4.45\% & +4.21\% & +4.47\% & +4.85\% & +1.89\% & +3.34\% & +3.01\% & +3.14\% & +\cellcolor[gray]{.7}\textbf{5.62}\% & +3.15\% \\ \hline
    \multirow{2}{*}{$K^{M}$}
                                     & RCV1/RCV2 & \cellcolor[gray]{.7}\textbf{+4.68}\% & +1.37\% & +0.77\% & +1.01\% & +2.60\% & +0.51\% & -0.25\% & +0.80\% & +0.25\%  & -- & -- \\ \cline{2-13}
                                     & JRC-Acquis & +4.85\% & +\cellcolor[gray]{.7}\textbf{5.89}\% & +4.29\% & +5.28\% & +5.15\% & +1.57\% & +3.57\% & +4.32\% & +4.69\% & +5.45\% & +4.11\% \\ \hline
  \end{tabular}
  }
  \caption{Average benefit (across languages
  $\lambda^{s}\in \mathcal{L}/\{\lambda^{t}\}$) obtained from the
  presence of $\lambda^{s}$ training data in classifying $\lambda^{t}$
  test data via \ftat. A greyed-out cell with a value in
  \textbf{boldface} indicates the language that has benefited most.}
  \label{tab:languagebenefit}
\end{table*}

\wasblue{A second conjecture we test} is if the benefit
$\beta(\lambda^{t})$ is negatively correlated with the accuracy of the
\textsc{Na\"ive} classifier for language $\lambda^{t}$ (once trained
with only 10\% of the $\lambda^{t}$ training examples, which is the
setting we have adopted in this section) as tested on $\lambda^{t}$
test data. This conjecture would also seem sensible, since we might
expect the benefit $\beta(\lambda^{t})$ to be higher when the
effectiveness of \textsc{Na\"ive} on language $\lambda^{t}$ is lower,
since in this case the margins of improvement are higher. In this case
too, the conjecture proves essentially correct, since the resulting
value of PCC is
$\rho(\beta(\lambda^{t}),F_{1}^{M}(\textsc{Na\"ive}(\lambda^{t})))=-0.605$
(p-val 0.08411), which indicates substantial negative correlation.

\subsection{Can we do without calibration?}
\label{sec:nocalibration}

\noindent As remarked in Section \ref{sec:funnelling}, one of the
aspects that contributes more substantially to the computational cost
of funnelling systems is probability calibration. The reason is that,
as also remarked in Section \ref{sec:learning}, calibration consists
in finding the optimal parameters of Equation \ref{eq:calibration}
through an extensive search within the space of parameter values. It
is thus of some interest to study whether we can do without
calibration at all, and what the effect of this would be. We have thus
run \ftat\ experiments in order to compare three alternative courses
of action:

\begin{enumerate}

\item \textsc{NoProb}: Renounce to converting classification scores
  into posterior probabilities. In this setting, a \ftat\ system is
  set up in which the meta-classifier (a) is trained with training
  documents represented by vectors $S(d_{l})$ of classification
  scores, and, (b) once trained, classifies documents represented by
  vectors $S(d_{u})$ of classification scores.

\item \textsc{NoCalib}: Convert classification scores into posterior
  probabilities, but renounce to calibrate them. This corresponds to
  employing a version of \ftat\ where, in place of \wasblue{the
  logistic function of Equation \ref{eq:calibration}, we use a
  non-parametric version} of it, which corresponds to Equation
  \ref{eq:calibration} with parameters $\alpha$ and $\beta$ fixed to 1
  and 0, respectively.

\item \textsc{Calib}: Employ the usual version of \ftat\ as defined in
  Section \ref{sec:twovariantsoffunnelling}.

\end{enumerate}

\noindent In Table \ref{tab:nocalibration}
\begin{table}[tb]
  \centering
  \begin{tabular}{|c|l||c|c|c|c|}
    \hline
    \multicolumn{2}{|c||}{\mbox{}}
    & \begin{sideways}\textsc{Na\"ive}\end{sideways} 
    & \begin{sideways}\textsc{NoProb}\end{sideways} 
    & \begin{sideways}\textsc{NoCalib}\phantom{0}\end{sideways} 
    & \begin{sideways}\textsc{Calib}\end{sideways} \\
    \hline
    \hline
    \multirow{2}{*}{$F_1^{\mu}$} 
    &	 RCV1/RCV2 &	 .776 $\pm$ .052 &	 .796\psddag  $\pm$ .045 &	 .789 $\pm$ .048 &	 \cellcolor[gray]{.7}\bf.802  $\pm$ .041\\     \cline{2-6}
    &	 JRC-Acquis &	 .559  $\pm$ .012&	 .585\sddag  $\pm$ .012 &	 .578 $\pm$ .012 &	 \cellcolor[gray]{.7}\bf.587  $\pm$ .009\\     \cline{2-6}    \hline    \hline
    \multirow{2}{*}{$F_1^{M}$}
    &	 RCV1/RCV2 &	 .467  $\pm$ .083&	 .463\psddag  $\pm$ .082 &	 .443 $\pm$ .086 &	 \cellcolor[gray]{.7}\bf.534  $\pm$ .066\\     \cline{2-6}
    &	 JRC-Acquis &	 .340  $\pm$ .017&	 .376\psddag  $\pm$ .021&	 .366 $\pm$ .015 &	 \cellcolor[gray]{.7}\bf.399  $\pm$ .013\\     \cline{2-6}    \hline    \hline
    \multirow{2}{*}{$K^{\mu}$} 
    &	 RCV1/RCV2 &	 .690  $\pm$ .074&	 .737\psddag  $\pm$ .062 &	 .716 $\pm$ .069 &	 \cellcolor[gray]{.7}\bf.760  $\pm$ .052\\     \cline{2-6}
    &	 JRC-Acquis &	 .429  $\pm$ .015&	 .478\sdpsdag  $\pm$ .018 &	 .465 $\pm$ .015 &	 \cellcolor[gray]{.7}\bf.490  $\pm$ .013\\     \cline{2-5}    \hline    \hline
    \multirow{2}{*}{$K^{M}$}
    &	 RCV1/RCV2 &	 .417  $\pm$ .090&	 .428\psddag  $\pm$ .087&	 .406 $\pm$ .091 &	 \cellcolor[gray]{.7}\bf.506  $\pm$ .073\\     \cline{2-6}
    &	 JRC-Acquis &	 .288  $\pm$ .016&	 .338\sdpsdag  $\pm$ .022 &	 .325 $\pm$ .016&	 \cellcolor[gray]{.7}\bf.365  $\pm$ .014\\     \cline{2-6}    \hline

  \end{tabular}
  \caption{Multilabel CLC results with alternative \ftat\
  settings. Notational conventions are as in Table
  \protect\ref{tab:mlmc}.  }
  \label{tab:nocalibration}
\end{table}
we report the results of running these three alternative systems; the
experimental setting is the same of Section \ref{sec:MLCLCresults},
and the results of Columns ``\textsc{Na\"ive}'' and ``\textsc{Calib}''
of Table \ref{tab:nocalibration} indeed coincide with those of Columns
``\textsc{Na\"ive}'' and ``\ftat'' of Table \ref{tab:mlmc}.

One fact that emerges from these results is that the standard
\textsc{Calib} setting always delivers the best performance, which is
unsurprising. A second fact that emerges is that the \textsc{NoCalib}
setting is always inferior to the \textsc{NoProb} setting. This is
surprising, since we might have conjectured \textsc{NoCalib} to
outperform \textsc{NoProb}, due to the fact that \textsc{NoCalib}
makes the outputs of the different base classifiers more comparable
among each other (by mapping them all into the [0,1] interval) than
the outputs used by \textsc{NoProb}; this finding \textit{de facto}
rules out \textsc{NoCalib} from further consideration.

Something that is much less clear, instead, is how \textsc{NoProb}
performs relative to \textsc{Na\"ive} and to the standard
\textsc{Calib} setting. In some cases \textsc{NoProb} performs very
well, almost indistinguishably from \textsc{Calib} (see $F_{1}^{\mu}$
results for JRC-Acquis), but in other cases it even performs worse
than the \textsc{Na\"ive} baseline, and dramatically worse than
\textsc{Calib} (see $F_{1}^{M}$ results for RCV1/RCV2).

All in all, these results confirm the theoretical intuition that
performing a full-blown probability calibration is by far the safest
option, and the one guaranteed to deliver the best results in all
situations.


\subsection{Efficiency}\label{sec:efficiency}

\noindent Table \ref{tab:efficiency}
\begin{table}[tb]
  \centering \resizebox{\textwidth}{!} {
  \begin{tabular}{|l||r|r|r|r|r|r|r|r|r|}
    \hline
    \multicolumn{1}{|c||}{\mbox{}} 
    & \multicolumn{1}{c|}{\begin{sideways}\textsc{Na\"ive}\end{sideways}} 
    & \multicolumn{1}{c|}{\begin{sideways}LRI\end{sideways}}
    & \multicolumn{1}{c|}{\begin{sideways}CLESA\end{sideways}} 
    & \multicolumn{1}{c|}{\begin{sideways}\wasblue{KCCA}\end{sideways}} 
    & \multicolumn{1}{c|}{\begin{sideways}DCI\end{sideways}} 
    & \multicolumn{1}{c|}{\begin{sideways}MLE\end{sideways}} 
    & \multicolumn{1}{c|}{\begin{sideways}\wasblue{MLE-LSTM}\end{sideways}} 
    & \multicolumn{1}{c|}{\begin{sideways}\fkfcv\phantom{0}\end{sideways}} 
    & \multicolumn{1}{c|}{\begin{sideways}\ftat\end{sideways}} \\
    \hline
    \hline
    \multirow{2}{*}{RCV1/RCV2} & 537 $\pm$ 69 &  5,506 $\pm$ 603 & 28,508 $\pm$5351 & \wasblue{18,204 $\pm$ 15} & 344 $\pm$ 51 & 1,293 $\pm$ 6 &  \wasblue{559 $\pm$ 103} & 1,041 $\pm$ 112 & \cellcolor[gray]{.7}\textbf{215} $\pm$ 16\\
    & 6 $\pm$ 0.3 & 91 $\pm$ 3 & 575 $\pm$ 10 & \wasblue{264 $\pm$ 7} & 9 $\pm$ 0.2 & 55 $\pm$ 1 &  \cellcolor[gray]{.7}\wasblue{\textbf{3} $\pm$ 0.1} & 13 $\pm$ 0.5 & 11 $\pm$ 0.4\\
    \hline
    \multirow{2}{*}{JRC-Acquis} & 6,005 $\pm$ 1,351 & 67,571 $\pm$ 2,070 & 63,497 $\pm$ 2,880 &\wasblue{57,563 $\pm$ 241} &  4,888 $\pm$ 1,136 & \cellcolor[gray]{.7}\textbf{4,435} $\pm$ 25 &  \wasblue{26,991 $\pm$ 915} & 13,127 $\pm$ 2,428 & 4,987 $\pm$ 208 \\ 
    & 84 $\pm$ 2 & 1,713 $\pm$ 6 & 4,049 $\pm$ 123 & \wasblue{1,372 $\pm$ 67} & 253 $\pm$ 3 & 874 $\pm$ 11 & \cellcolor[gray]{.7}\wasblue{\textbf{6} $\pm$ 0.4} & 312 $\pm$ 4 & 278 $\pm$ 2\\
    \hline
  \end{tabular}
  }
  \caption{Computation times (in seconds); 1st rows indicate training
  times while 2nd rows report testing times.}
  \label{tab:efficiency}
\end{table}
reports training times and testing times for all the methods discussed
in this paper, as clocked on our two datasets; each reported value is
the average value across the 10 random trials. The experiments were
run on a machine equipped with a 12-core processor Intel Core i7-4930K
at 3.40GHz with 32 GB of RAM under Ubuntu 16.04 (LTS).  \wasblue{For
\textsc{MLE-LSTM}, the times reported correspond to our Keras
implementation running on a Nvidia GeForce GTX 1080 equipped with 8 GB
of RAM.}
We limit our analysis to the multilabel CLC setup of Section
\ref{sec:MLCLCresults} (thus skipping the discussion of the setups of
Sections \ref{sec:monolingualresults} and \ref{sec:lessresourced}) (a)
since multilabel CLC is the most interesting context, and (b) since
for the setups discussed in Sections \ref{sec:monolingualresults} and
\ref{sec:lessresourced} we have run only \ftat\ and \textsc{Na\"ive}.

The most interesting fact that emerges from Table \ref{tab:efficiency}
is that the superior accuracy of \ftat\ does not come at a
price. Indeed, \ftat\ often turns out to be one of the most efficient,
or sometimes \emph{the} most efficient, among the methods we test; in
particular, both at training time and testing time it is one order of
magnitude faster than LRI, its most important competitor. \fkfcv\ is,
as previously observed, much more expensive to train than \ftat, due
to the much higher number of training and probability calibration
rounds that it requires. \textsc{CLESA} is clearly the most
inefficient of all methods, which is explained by the fact that each
(labelled or unlabelled) document requires one document similarity
computation for each feature in its vectorial representation. The
higher training-time efficiency of \ftat\ with respect to
\textsc{Na\"ive} is certainly also due to the fact that, as mentioned
in Section \ref{sec:learning}, we do not perform any optimization of
the $C$ parameter for the base classifiers of \ftat, while we do for
the classifiers of \textsc{Na\"ive}; should we perform this parameter
optimization the computational cost of \ftat\ would certainly
increase, but so probably would also the differential in effectiveness
between \ftat\ and all the other baselines.

\wasblue{Note that the most efficient method in testing mode is
\textsc{MLE-LSTM}, especially in the case of JRC-Acquis, where it is
one order of magnitude faster than the 2nd fastest method
(\textsc{Na\"ive}). The reasons are twofold: (a) as noted above, the
MLE-LSTM experiments have been run on hardware different from the
hardware used for all the other experiments, so comparisons are
difficult to make; (b) in models trained via deep learning, such as
MLE-LSTM, testing reduces to a simple forward pass through the network
connections, something which can be performed very quickly by
exploiting the massive parallelism offered by modern GPUs.}

\section{Can funnelling be used in the zero-shot setting?}
\label{sec:ZSCLC}

\noindent The experiments we have discussed so far have assumed a
setting in which there is a \wasblue{non-zero} number of training
examples for each of the target languages, and in which the training
examples for the source languages have thus the goal of
\emph{improving} the accuracy of the classifiers generated from the
training examples of the target languages.  We might wonder whether
funnelling can also be used in a zero-shot setting, i.e., one in which
there are no training examples for the target languages, and in which
the training examples for the source languages would have the goal of
allowing to generate classifiers for the target languages that could
otherwise not be generated at all.

Unfortunately, the answer is no.  To see why, for simplicity let us
discuss \ftat\ (the case of \fkfcv\ is analogous). If there are no
positive training documents for pair $(\lambda_{i},c_{j})$, this means
that (as noted in Section \ref{sec:learning}) the base classifier
$h^{1}_{i}$ generated from the negative examples only (i.e., from the
examples in $\lambda_{i}$ that are positive for some other class in
$\mathcal{C}/\{c_{j}\}$) is a trivial rejector for $c_{j}$, i.e., one
that only returns scores $h^{1}_{i}(d_{u},c_{j})=0$ for all unlabeled
documents $d_{u}$ written in language $\lambda_{i}$. By definition,
the calibration function turns all these scores into posterior
probabilities $\Pr(c_{j}|d_{u})=0$. As a result, when the negative
training examples are reclassified by $h^{1}_{i}$ for generating
vectorial representations that contribute to training the
meta-classifier, these negative training examples originate vectors
that contain a 0 for class $c_{j}$. Since these are all negative
examples, the meta-classifier is trained to interpret a value of 0 in
the vector position corresponding to $c_{j}$ as a perfect predictor
that the document does not belong to $c_{j}$. As a result, when an
unlabelled document in language $\lambda_{i}$ is classified, the base
classifier returns a value $h^{1}_{i}(d_{u},c_{j})=0$, which is
converted into a posterior probability $\Pr(c_{j}|d_{u})=0$, which is
thus interpreted as unequivocally indicating that $d_{u}$ does not
belong to $c_{j}$, independently of the contributions coming from
classes other than $c_{j}$ and languages other than $\lambda_{i}$. The
entire 2-tier
classifier is then a trivial rejector for pair
$(\lambda_{i},c_{j})$.\footnote{Note that this is confirmed by the
experiments plotted in Figure \ref{fig:relimp}, where for $x=0$ it
holds that $F_{1}^{M}=0$ for all languages $\lambda_{i}$. In fact,
when there are no training examples for the target language ($x=0$)
the entire 2-tier classifier is, as observed above, a trivial
rejector, which means that $TP$ is 0 and, as a consequence, $F_{1}$ is
0 too, as clearly visible for all plots in the figure.} This shows
that funnelling is unsuitable for dealing with the scenario in which
there are no training examples for the target languages.

\wasblue{This problem has prompted us to devise ways of enabling
funnelling to also operate in ``zero-shot mode'' (i.e., on documents
expressed in languages for which no training documents are available).
The basic idea is to add a ``zero-shot classifier''
$h^{1}_{(|\mathcal{L}|+1)}$ (which for notational simplicity we denote
by $h^{1}_{z}$) to the 1st-tier classifiers, i.e., a classifier that
is to be invoked whenever a document written in any language different
from the ones in $\mathcal{L}$ (i.e., from the languages for which
training examples do exist) needs to be classified. This means that
the 2nd-tier classifier is trained also on (and also receives as
input) the posterior probabilities returned by $h^{1}_{z}$, which thus
needs to be a well calibrated classifier.  Note that this modification
fits smoothly into the framework, since funnelling makes very few
assumptions about the characteristics of the base classifiers.  For
simplicity, we here derive the adaptation for \ftat; the case of
\fkfcv\ is similar.

More formally, let $\mathcal{L}$ be a set of languages for which
labelled training examples are available.
In this new variant of the funnelling system, in the 1st tier there
are (as usual) $|\mathcal{L}|$ language-specific classifiers
$h_1^1,\ldots,h_{|\mathcal{L}|}^1$,
plus one classifier $h^{1}_{z}$ trained (according to some method yet
to be specified) on
\emph{all} the training examples in any of the languages in
$\mathcal{L}$.  For each training document $d_l$ in language
$\lambda_i$, \emph{two} 
vectorial representations are generated that are used in training the
2nd-tier classifier $h^2$, i.e., the vector of posterior probabilities
$$(f_i(h^1_i(d_l,c_1)),\ldots,f_i(h^1_i(d_l,c_{|\mathcal{C}|})))$$ from
the language-dependent classifier $h_i^1$, and the vector of posterior
probabilities
$$(f_{z}(h^1_z(d_l,c_1)),\ldots,f_z(h^1_z(d_l,c_{|\mathcal{C}|})))$$ from
the zero-shot classifier $h^1_z$.
Therefore, $h^2$ is trained on twice the number of
$|\mathcal{C}|$-dimensional vectors with respect to the one we
considered in the previous sections.

When a new unlabelled document $d_u$ expressed in language $\lambda$
is submitted for classification, two scenarios are possible:

\begin{enumerate}

\item $\lambda\in\mathcal{L}$: this case reduces to funnelling as
  discussed in the previous sections, that is, (a) the document is
  first represented in its corresponding language-specific feature
  space,
  (b) a vector of posterior probabilities is then obtained using the
  corresponding language-specific 1st-tier classifier, and (c) the
  2nd-tier classifier $h^2$ takes the final decision;

\item $\lambda\notin\mathcal{L}$: in this case, (a) the document is
  first represented in the feature space of $h^1_z$,
  (b) a vector of posterior probabilities is then obtained using the
  calibrated 1st-tier classifier $h^1_z$, and (c) the 2nd-tier
  classifier $h^2$ takes the final decision.

\end{enumerate}

\noindent \textsc{CLESA}, \textsc{MLE}, and \textsc{MLE-LSTM} are
possible methods by means of which the representations $\phi^1_z(d)$
in the feature space of $h^1_z$ can be obtained. For example,
\textsc{MLE} trains a classifier on representations of the documents
consisting of averages of multilingual word embeddings.  Since
multilingual word embeddings are aligned across languages
\cite{Conneau:2018bv}, the same classifier would, in principle, be
capable of classifying a document written in any language $\lambda$
(possibly with $\lambda\notin\mathcal{L}$) for which pre-trained and
aligned word embeddings are available.  Similar considerations enable
\textsc{CLESA} to work
with documents in languages not in $\mathcal{L}$, as long as a set of
comparable Wikipedia articles are available for their language.}

\wasblue{For our experiments we choose \textsc{MLE} as the method to
generate the 1st-tier zero-shot classifier, because of the good
trade-off between effectiveness and efficiency it has shown in our
previous experiments.  We call the resulting ZSCLC classification
method \ftat-\textsc{MLE}.}

\wasblue{In order to test \ftat-\textsc{MLE} we run experiments in
which we \wasblue{incrementally} augment the set of languages for which
training examples are available.  In each new experiment, the training
set of a new language is added, while the languages for which training
data have not been added yet are dealt with by the zero-shot
classifier.  For example, after the third experiment, the training
data for the three languages \{DA,DE,EN\} \wasblue{(i.e., Danish, German,
English)} have been added to the training set (we add languages
following the alphabetical order). The test set is instead fixed, and
always contains all test examples of all languages.}

\begin{figure}[tb]
  \centering \includegraphics[width=\textwidth]{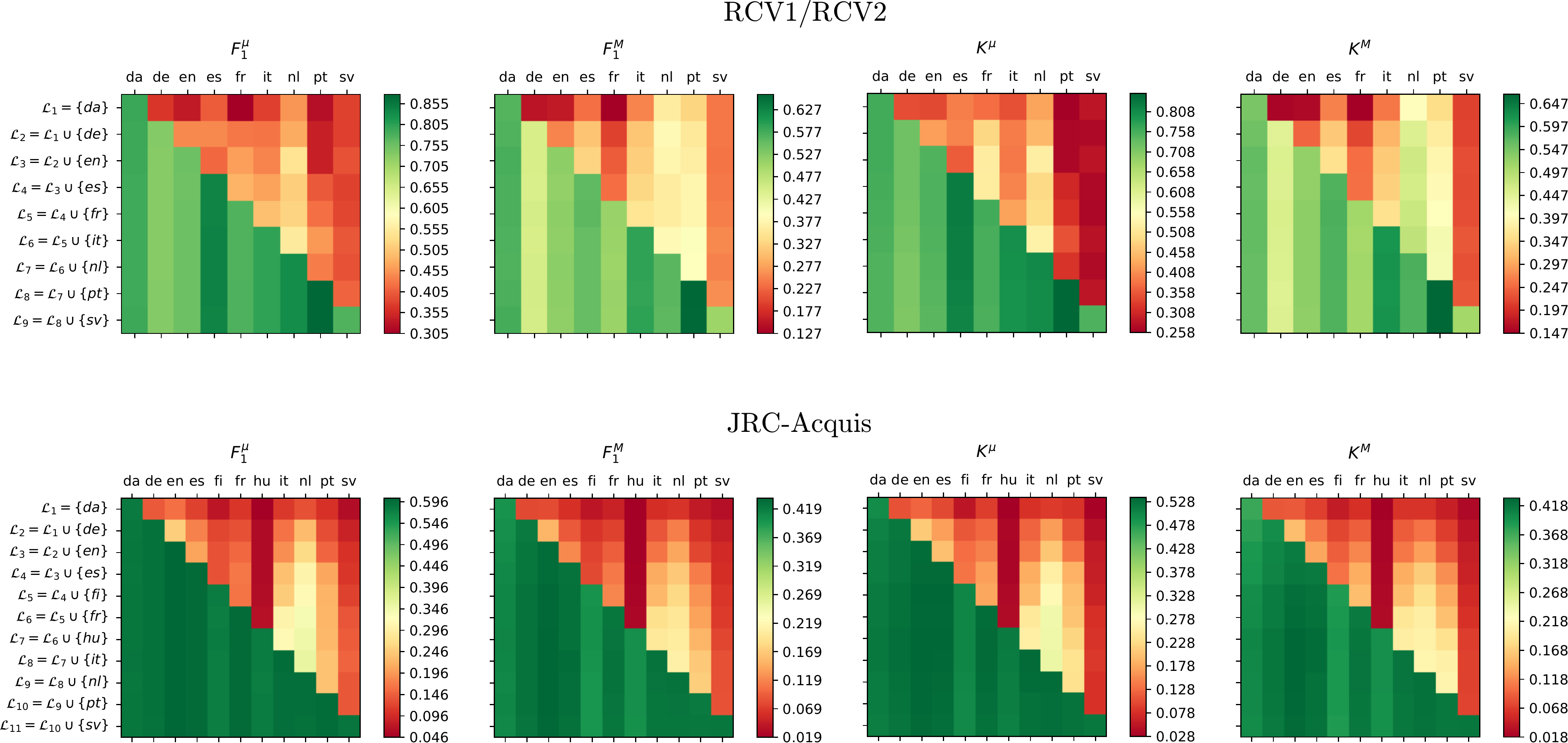}
  \caption{\wasblue{Zero-shot CLC experiments using \ftat-\textsc{MLE}
  in RCV1/RCV2 (top) and JRC-Acquis (bottom) for the four evaluation
  measures (from left to right) $F_1^{\mu}$, $F_1^{M}$, $K^{\mu}$, and
  $K^{M}$. In each square matrix, columns represent test languages,
  while rows represent training sets with an increasing (from top to
  bottom) number of languages.}}
  \label{fig:zeroshot}
\end{figure}

\wasblue{The results of our experiments are displayed in Figure
\ref{fig:zeroshot}, where colours are used instead of numerical data
in order to make patterns and trends more evident.  Each of the 8
square matrixes of coloured cells represents the experiments performed
on one of our 2 datasets and using one of our 4 evaluation measures;
each cell in a matrix represents the accuracy obtained using the
training data for a given group of languages (indicated on the row)
and the test data for a given language (indicated on the column). In
each such matrix, the lower triangular matrix reflects the
classification outcomes on test languages which are represented in the
training set; because of this, accuracy results are typically high
(green). The upper triangular matrix represents the outcomes for
languages that are \emph{not} represented in the training data, which
thus tend to obtain lower scores (red).}  \wasblue{For the sake of
visualization we have individully normalized each of the 8 colour
maps, i.e., each such map contains both a dark green
cell 
and a dark red cell, corresponding to the highest value and lowest
value of the evaluation function for that colour map, respectively
(i.e., colours have a relative meaning, and not an absolute one).}

\wasblue{One clear pattern that emerges from Figure \ref{fig:zeroshot}
is that the piecemeal addition of languages to the training set
improves the classification accuracy for the yet unseen (i.e., not
represented in the training set) languages, as witnessed by the
gradual change in colour through columns, from dark red on top to
lighter red towards the bottom.

Notwithstanding this, a similar improvement does not clearly emerge
for the already seen languages, i.e., the addition of languages to the
training set does not seem to boost the classification accuracy for
the languages already represented in the training set.  However, such
an improvement does exist in the ``pure'' version of \ftat, as
verified and discussed in Sections \ref{sec:monolingualresults},
\ref{sec:lessresourced}, and \ref{sec:languages}.

A possible explanation for this anomaly might be a negative
side-effect introduced by the $h^1_z$ classifier into the
meta-classifier.
\wasblue{The reason is that the metaclassifier is fed with posterior
probabilities generated by classifiers working with differently
characterized data.  Inasmuch as the calibration process fails to
deliver perfectly calibrated probabilities, the two representations
might happen to be distributed differently, thus causing an
``interference'' effect between the two sources of information.  A
possible solution to this problem, that we plan to investigate in
future research, is to reduce (along with \cite{ganin2016domain}) the
gap between the two distributions via adversarial training, i.e., to
learn a transformation of the posterior probabilities from one
distribution that 
makes them indistinguishable from the posterior probabilities of the
other distribution, as judged by a discriminator model concurrently
trained to discriminate between the two distributions.}

For the moment being, the experiments discussed in this section seem
to indicate (a) that funnelling, as a framework, can indeed be adapted
to ZSCLC, but (b) that better ways of combining the posterior
probabilities returned by the 1st-tier classifiers should be
investigated for ZSCLC. This is something we plan to do in future
research.}


\section{Conclusion}\label{sec:conclusion}

\noindent This paper presents (a) a novel 2-tiered ensemble learning
method for heterogeneous data, and (b) the first (to the best of our
knowledge) application of an ensemble learning method to multilingual
(and more specifically: cross-lingual multilabel) text
classification. While similar to stacked generalization, this ensemble
learning method (that we dub ``funnelling'') is different from it
because the base classifiers are specialized, each catering for a
different type of objects characterized by its own feature space.  In
cross-lingual classification, this means that different base
classifiers deal with documents written in different languages;
funnelling makes it possible to bring them all together, so that the
training examples for all languages in $\mathcal{L}$ contribute to the
classification of all unlabelled documents, irrespectively of the
language $\lambda\in\mathcal{L}$ they are written in.

One advantage of funnelling is that it is learner-independent; while
in this paper we test it with SVMs as the learning method, it can be
set up to use (a) any learning device that outputs non-binary
classification scores (for the base classifiers), and (b) any
learning device that accepts numeric feature values as input (for the
meta-classifier).
An additional advantage of funnelling is that, unlike several other
multilingual methods, it does not require external resources, either
in the form of multilingual dictionaries, or machine translation
services, or external parallel corpora.

The extensive experiments we have run on a comparable 9-language
corpus (RCV1/RCV2) and on a parallel 11-language corpus (JRC-Acquis)
against a number of state-of-the-art baseline methods, show that
\ftat\ (the better of two funnelling methods we have tested) (a)
almost always outperforms all baselines, irrespectively of evaluation
measure, averaging method, and dataset; (b) delivers improvements
over the na\"ive monolingual baseline more consistently (i.e., for all
tested languages, datasets, evaluation measures, averaging methods)
than any other baseline considered; and (c) is among the most
efficient tested methods, at both training time and testing time. All
this has been confirmed across a range of experimental settings, i.e.,
binary or multilabel,
monolingual or cross-lingual. The two main factors behind the success
of funnelling in cross-lingual multilabel classification are (a) its
ability to leverage the training examples written in any language in
order to classify unlabelled examples written in any language, and
(b) its ability to leverage the stochastic dependencies between
different classes.


Funnelling is useful whenever (a) the data to be classified comes in
different types that require different feature representations,
\emph{and} (b) despite these differences in nature, all data need to
be classified under a common classification scheme $\mathcal{C}$. We
are currently testing funnelling in other such contexts, e.g.,
classifying images of products and textual descriptions of products
under the same set $\mathcal{C}$ of product classes.



\begin{acks}
The present work has been supported by the \textsf{ARIADNEplus} project, funded by the European Commission (Grant 823914) under the H2020 Programme INFRAIA-2018-1. The authors' opinions do not necessarily reflect those of the European Commission.
\end{acks}


\bibliographystyle{ACM-Reference-Format}
\bibliography{Funnelling}

\end{document}